\documentclass[lettersize,journal]{IEEEtran}
\IEEEoverridecommandlockouts   

\usepackage{fancyhdr}
\fancypagestyle{firstpage}{
  \fancyhf{}
  
  \fancyfoot[L]{\footnotesize © 2024 IEEE. Personal use of this material is permitted. Permission from IEEE must be obtained for all other uses, in any current or future media, including reprinting/republishing this material for advertising or promotional purposes, creating new collective works, for resale or redistribution to servers or lists, or reuse of any copyrighted component of this work in other works.}
}

\usepackage{cite}
\usepackage{amsmath,amssymb,amsfonts}
\usepackage{graphicx}
\usepackage{textcomp}
\usepackage{xcolor}
\usepackage{dsfont}
\usepackage{subfig}
\usepackage{float}
\usepackage{bbm}

\usepackage{scalefnt}
\usepackage{tikzscale}
\usepackage{tikz}
\usepackage{pgfplots}
\usepackage{lipsum,adjustbox}
\pgfplotsset{compat=1.17}
\usepackage{changepage}
\usetikzlibrary{shapes.geometric}
\usetikzlibrary{calc}
\usetikzlibrary{positioning}
\usetikzlibrary{intersections}
\usetikzlibrary{decorations.fractals,spy}

\usepackage{acronym}

\usepackage{algorithm}
\usepackage{algpseudocode}

\usepackage{url}

\usepackage{siunitx}

\usepackage[normalem]{ulem}

\usepackage{makecell}

\newtheorem{rem}{Remark}

\DeclareMathOperator*{\minimize}{\textbf{minimize}}
\DeclareMathOperator{\E}{\mathbb{E}}
\DeclareMathOperator{\stt}{\textbf{subj. to}}
\DeclareMathOperator*{\argmin}{\textbf{argmin}}

\definecolor{responsecolor}{rgb}{0,0,0}

\newcommand{\DCAUKS}{{\textrm{D}^2\textrm{C}^2\textrm{-AUKS}}}

\newacro{AD}{Autonomous Driving}
\newacro{NMPC}{Nonlinear Model Predictive Control}
\newacro{RTNMPC}{Real-Time NMPC}
\newacro{xDT}{Executable Digital Twin}
\newacro{MiL}{Model-in-the-Loop}
\newacro{HiL}{Hardware-in-the-Loop}
\newacro{ViL}{Vehicle-in-the-Loop}
\newacro{TiL}{Twin-in-the-Loop}
\newacro{BO}{Bayesian Optimization}
\newacro{EKF}{Extended Kalman Filter}
\newacro{UKF}{Unscented Kalman Filter}
\newacro{DR}{Domain Randomization}
\newacro{DA}{Domain Adaptation}
\newacro{FMU}{Functional Mock-up Unit}
\newacro{DT}{Digital Twin}
\newacro{IUKF}{Iterative UKF}
\newacro{IEKF}{Iterative EKF}
\newacro{UT}{Unscented Transform}
\newacro{RMS}{Root-Mean-Square}
\newacro{SOCSE}{Spectral Orthogonal Collocation with Safety Envelope}
\newacro{OCP}{Optimal Control Problem}
\newacro{NLP}{NonLinear Programming}
\newacro{KPI}{Key Performance Indicator}

\def\BibTeX{{\rm B\kern-.05em{\sc i\kern-.025em b}\kern-.08em
    T\kern-.1667em\lower.7ex\hbox{E}\kern-.125emX}}
\begin{document}

\title{Learning Based NMPC Adaptation for Autonomous Driving using Parallelized Digital Twin 
}

\author{{Jean Pierre Allamaa$\,\,^\textrm{1,2}$, Panagiotis Patrinos$\,\,^\textrm{2}$, Herman Van der Auweraer$\,\,^\textrm{1,3}$, Tong Duy Son$\,\,^\textrm{1}$}
\thanks{$^\textrm{1}$  Siemens Digital Industries Software,  3001 Leuven, Belgium. Email: \tt \{jean.pierre.allamaa, herman.van-der-auweraer.ext@siemens.com, son.tong\}@siemens.com}
\thanks{$^\textrm{2}$  Dept. Electr. Eng. (ESAT) - STADIUS research group, KU Leuven, 3001 Leuven, Belgium. Email: \tt panos.patrinos@esat.kuleuven.be}
\thanks{$^\textrm{3}$  Dept. Mechanical Eng. - LMSD research group, KU Leuven, 3001 Leuven, Belgium}
}

\maketitle
\thispagestyle{firstpage}

\begin{abstract}
	In this work, we focus on the challenge of transferring an autonomous driving controller from simulation to the real world (i.e. Sim2Real). We propose a data-efficient method for online and on-the-fly adaptation of parametrizable control architectures such that the target closed-loop performance is optimized while accounting for uncertainties as model mismatches, changes in the environment, and task variations. The novelty of the approach resides in leveraging black-box optimization enabled by Executable Digital Twins (xDTs) for data-driven parameter calibration through derivative-free methods to directly adapt the controller in real-time. The \ac{xDT}s are augmented with Domain Randomization for robustness and allow for safe parameter exploration. The proposed method requires a minimal amount of interaction with the real-world as it pushes the exploration towards the \ac{xDT}s. We validate our approach through real-world experiments, demonstrating its effectiveness in transferring and fine-tuning a \ac{NMPC} with 9 parameters, in under 10 minutes. This eliminates the need for hours-long manual tuning and lengthy machine learning training and data collection phases. Our results show that the online adapted \ac{NMPC} directly compensates for the Sim2Real gap and avoids overtuning in simulation. Importantly, a 75\% improvement in tracking performance is achieved and the Sim2Real gap over the target performance is reduced from a factor of 876 to 1.033.
\end{abstract}%

\acresetall 

\section{Introduction}
\begin{figure*}[h!]
	\centering
	\begin{adjustbox}{width=\textwidth}
		\input{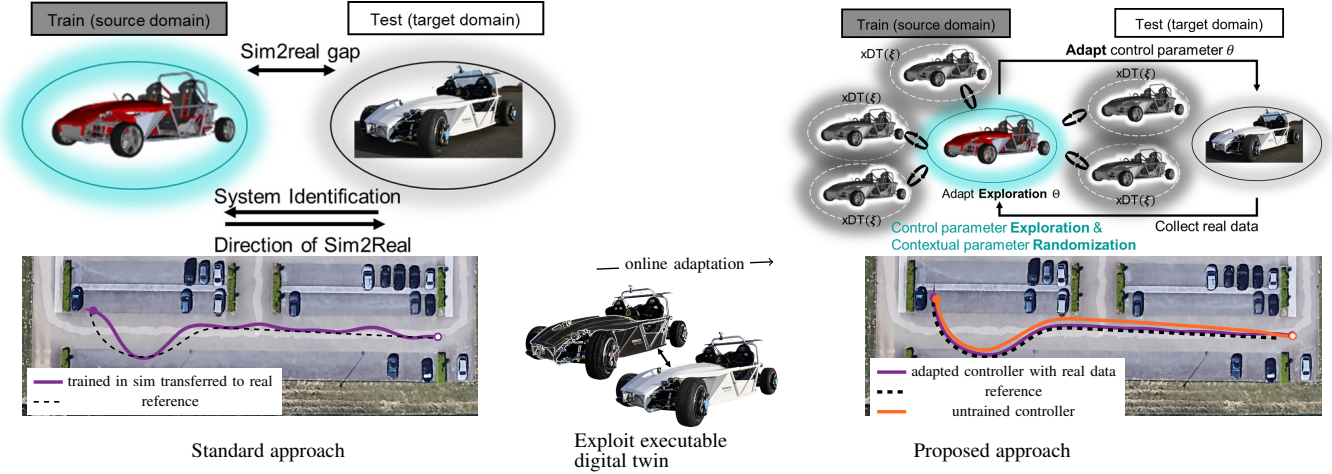}
	\end{adjustbox}
	\caption{Vehicle-in-the-loop performance. Left: control policy over trained in simulation with only the nominal identified vehicle model to achieve less than 15 cm of accuracy, fails to transfer to the real-world and results in an unsafe driving style. \textcolor{responsecolor}{Right: Enhancing the Sim2Real transfer through data-driven controller adaptation based on domain randomization, domain adaptation, high-fidelity simulation and with fine-tuning from real-world data}} 
	\label{fig:sim2real_vil_prob}
\end{figure*}

\IEEEPARstart{P}{erformance} of advanced control strategies often depends on the choice of (hyper)parameters. A powerful control strategy, such as \ac{NMPC}, is challenged in the context of verification and validation because many parameters need to be tuned simultaneously. This tuning process requires engineering knowledge and is often difficult to scale to new applications, new environments, and new tasks. Although heuristics and methods exist for tuning some controllers, researchers often face a second barrier when proceeding to deployment, i.e., the simulation to reality transfer: Sim2Real is a commonly faced challenge as controllers tuned in one domain (simulation) fail to transfer to a target domain (real world), and result in deceiving performance as in Figure~\ref{fig:sim2real_vil_prob} due to domain shifts, noise and uncertainties. This raises the need to close the loop between prior expectation from simulation, with actual feedback from the real world, in the form of online adaptation to encapsulate the true uncertainties and Sim2Real gap. To tackle these challenges, a strategy with an adaptation layer learns to overcome these errors without overestimating them, by directly optimizing the target domain performance. \textcolor{responsecolor}{This contrasts with traditional approaches in the automotive industry that rely on a one directional V-cycle for verification and validation. The standard V-cycle goes from \ac{MiL} to \ac{HiL} and finally to \ac{ViL}. During \ac{ViL}, the performance of the full closed-loop system is determined, and the engineer is faced with two options: tedious online calibration for end-of-line tuning or iterating back to the \ac{MiL} phase. Therefore, we aim to bridge the gap between simulation and reality by enabling a bidirectional communication between them so that the simulation runs in parallel with the real counterpart. This allows for safe and cheap exploration of the parameter space in simulation, which can be exploited by the real counterpart to calibrate for the optimal parameters.}

Several research directions exist, focusing on including learning in the control strategy. Reinforcement and imitation learning are two examples of those strategies used to either train a control policy from scratch, which can be unsafe for safety-critical systems such as in autonomous vehicles, or to tune a predefined safe controller such as \ac{NMPC}. Although these methods showed great success and are scalable as in~\cite{Acerbo2022EvaluationOM}, they are often data hungry, require long training time and assume that numerical gradient methods could be employed given differentiability between the output and the parameters. Another direction for incorporating learning and adaptation in the \ac{NMPC} is through online learning. 
In~\cite{an_adaptive_2009}, the authors propose to optimize over the weighting matrices on the slack variables of the MPC's terminal constraints to improve robustness in specified critical zones. The authors of~\cite{frohlich_contextual_2022} propose a parameter tuning approach based on contextual \ac{BO}. In the work of~\cite{nobar2024guided}, a novel BO approach is used for automatic control tuning by offloading exploration towards the \ac{DT}. Moreover, a popular method for adapting the \ac{NMPC} model is through residual dynamics learning using neural networks~\cite{jiahao_online_2022} or Gaussian Processes~\cite{hewing_learning-based_2020}. Adaptation to reduce conservatism compared to robust control as in~\cite{8618694}, and safe environment learning~\cite{schuurmans_GeneralFrameworkLearningBased_2023}, are also relevant. However, most of these methods affect the real-time applicability and structure of the \ac{NMPC} or require a large amount of data to train offline. Notably, some of those methods like vanilla BO are not suitable for online training as parameter adaptation requires exploring the parameter system on the target system. Furthermore, data-driven parameter tuning in the form of classical adaptive control has shown promising results~\cite{MARINO199755} and was applied to tuning MPC~\cite{8594267}. However, most of these approaches are applied solely to linearized systems and for simple tracking problems. Therefore, they are limited in their scalability and region of validity due to potential oversimplification. Although several previous works have tackled the parameter learning aspect, most of the results were either limited to simulation or controlled lab environment. 

This paper extends the previous work of~\cite{ALLAMAA2022385} \textcolor{responsecolor}{by including experimental validation of the automatic tuning for a real-time NMPC on a real road vehicle, alongside parallelization of \ac{DT}s through an industrial standard known as FMU, and providing necessary convergence conditions for the algorithm}.

To address the previously identified gaps, we present Data-Driven Controller Calibration with Adaptive Unscented Kalman filter and SPSA ($\DCAUKS$): a method that offers a fast and on-the-fly automatic controller adaptation. It employs gradient-free methods for stochastic optimization that enable learning the controller's optimal parameters in a limited amount of time with sample and data efficiency. Notably, the method avoids trial-and-error in the parameter search and allows parallelization unlike \ac{BO}, rendering it efficient for controller tuning on real-systems. A key aspect of the work is to optimize over possibly non-differentiable and non-analytical performance measures by means of parameters estimation using \ac{UKF} as first presented in~\cite{menner2023automated}. Unlike~\cite{menner2023automated}, we include \ac{DR} for robustness and more accurate parameter distribution, a sampling-based gradient estimation known as SPSA~\cite{Spall1998ANOO} to speed up the tuning process, and offload online exploration towards \ac{xDT}s. Moreover, we extend the method in~\cite{ALLAMAA2022385} further by carefully adapting the noise covariance matrices to provide sufficient conditions for convergence and apply the entire framework to a real vehicle\footnote{Abstract and experiments video at \url{https://youtu.be/62PgNHciIJA}.}. As the method optimizes for the closed-loop performance without changing the structure of the \ac{NMPC}, it is suitable for real-time applications. In the adaptation framework, we exploit the \ac{xDT}s to explore different parameter combinations efficiently and rapidly, in a digital yet highly reliable environment. This permits to offload the exploration burden towards the \ac{xDT}s, eliminating the need for costly and suboptimal manual tuning. The main contributions are:%
\begin{itemize}%
	\item a fast and safe online adaptation framework that requires little interaction with the environment and is able to directly compensate for uncertainties;
	\item A parallelization approach for executable digital twins in the loop;
	\item Parameter tuning through stochastic and black-box optimization that is robust to noise through adaptive Kalman filter covariance matrices;
	\item Experimental validation of the method on a real-world vehicle with a \ac{RTNMPC} for end-of-line tuning, resulting in 75\% tracking performance improvement within few iterations and a drop in the Sim2Real gap from a factor of 876 to 1.033.
\end{itemize}
The paper is structured as follows: Section~\ref{section:prelim_sim2real} provides a background on relevant Sim2Real techniques and introduces the proposed parallelization of multiple \ac{xDT}s.
Section~\ref{section:learning_framework} presents the learning and adaptation framework through \ac{xDT} and derivative-free optimization (DFO). Furthermore, section~\ref{section:adaptation_simulation} validates the method in simulation for an autonomous valet parking application and shows the benefit of domain randomization. Moreover, we present and discuss the results of the experimental validation of the automatic \ac{NMPC} calibration on the road vehicle in section~\ref{section:adaptation_experimental}. Finally, the limitations of the work are presented in section~\ref{section:limitations} before concluding the paper in section~\ref{section:conclusion}.
\section{Preliminary on Sim2Real Techniques}
\label{section:prelim_sim2real}
Manual tuning is time-consuming, costly, and can lead to performance discrepancies among different products. Moreover, transferring controllers from simulation to the real-world is challenging due to changing environments and a domain shift known as Sim2Real gap. Automatic tuning and continuous data-driven adaptation is a promising solution to these issues as in~\cite{twin_in_the_loop_23, nobar2024guided}. In this paper, we propose a black-box optimization-driven adaptation framework specifically designed to find suitable control parameters for the \ac{NMPC} by deploying \ac{xDT}s in parallel with the real vehicle as depicted in Figure~\ref{fig:general_adaptation_framework}. We first provide an overview of existing Sim2Real strategies and then introduce the concept of \ac{xDT}s and highlight their parallelization capability through a standalone library instance.

\subsection{Sim2Real through Domain randomization and adaptation}
We address the challenge of transferring a controller designed in a source domain to a target domain. This concerns for example the transfer of a learned control policy from a training plant to a testing plant with different parameters. 
Specifically, we tackle the Sim2Real transfer which suffers from Sim2Real gap curse. Tuning parameters solely in the source domain poses a significant risk of ineffective transfer due to domain or model mismatches, uncertainties, dynamic environmental changes, and lack of robustness. Therefore, additional techniques to recover or limit the performance loss in the target domain are necessary.
Two common methods to address this issue are \ac{DR} and \ac{DA}. \ac{DR} involves injecting uncertainties into the test domain by introducing noise and model mismatches (e.g. mass, inertia, friction coefficients, varying tasks) to enhance the robustness of the learned policy against potential disturbances. \ac{DR} has been widely used in the field of robotics and autonomy~\cite{peng_sim_real_2018,Voogd2022ReinforcementLF}. On the other hand, \ac{DA} involves training in a latent space using a domain invariant feature representation. An example of learning in higher-dimensional spaces by automatic domain adaptation to facilitate transfer is presented in~\cite{Frhlich2019BayesianOF}. A survey on Sim2Real techniques can be found in~\cite{sim2real_10242366}.

\subsection{xDT and FMU parallelization}
Simulations play a vital role in control applications in the development and validation processes.  As safety concerns and cost per simulation are minimal, researchers can explore various scenarios and environments using a \ac{MiL} approach. When the algorithms are safe and robust, they are transitioned to \ac{HiL} and \ac{ViL}. The standard flow consists of first identifying an accurate high-fidelity model, then using it to train or validate the policies in close-loop as depicted in the left framework in Figure~\ref{fig:sim2real_vil_prob}. Although successful in many applications, this standard approach fails to transfer between domains and requires fine-tuning inspired by engineering expertise. Particularly, performance degradation occurs due to the simulation-optimization bias~\cite{DBLP:journals/pami/MuratoreG021}. With the emergence of automated vehicles, this involves a considerable end-of-line tuning to recover the lost performance because of the Sim2Real gap. To address this, the concept \ac{xDT}s has been proposed~\cite{Hartmann2022TheED}: a high-fidelity twin model of the plant that can be instantiated to run in parallel with its real counterpart, on embedded hardware. The \ac{xDT} captures the complex dynamics, and allows the exchange of data between the digital and real twins. Moreover, it is also possible to instantiate not one, but several \ac{xDT}s to allow for real-time testing of multiple configurations while safely and efficiently exploring in the digital world. \textcolor{responsecolor}{This concept has been further elaborated as a \ac{TiL} in~\cite{twin_in_the_loop_23} to automatically tune a compensator for the Sim2Real gap using \ac{BO}, showing the benefit of employing a DT on-the-go for better prediction.}

\begin{figure}
	\centering
	\includegraphics[width=0.45\textwidth]{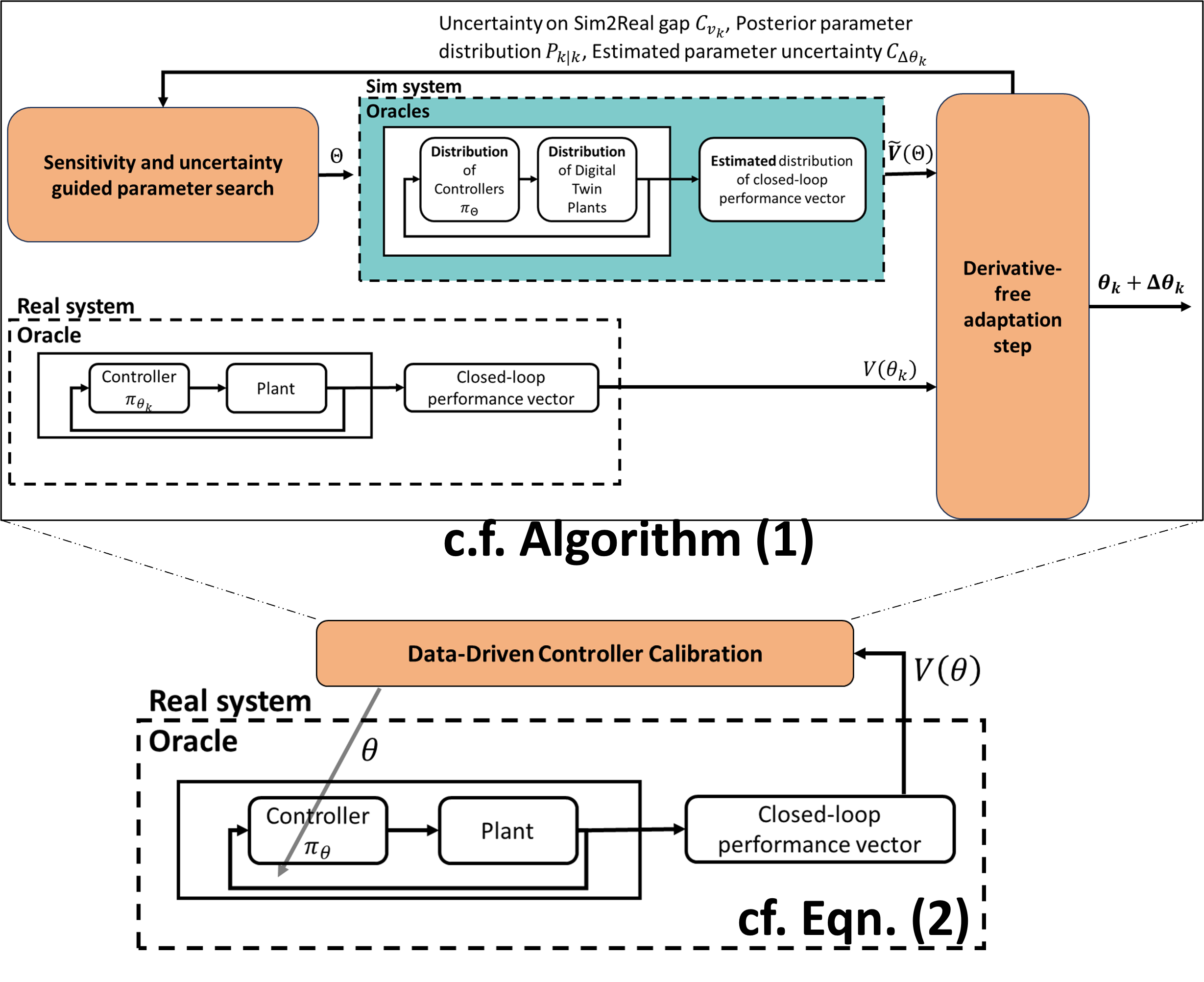}
	\caption{\textcolor{responsecolor}{$\DCAUKS$: on-the-go parametrizable controller calibration that offloads the burden of parameter search and sensitivity analysis towards multiple Digital Twins, and efficiently closes the Sim2Real gap through real-world data flow}}
	\label{fig:general_adaptation_framework}
\end{figure}

\begin{figure*}[h!]
	\centering
	\begin{adjustbox}{width=0.75\textwidth}
		\includegraphics{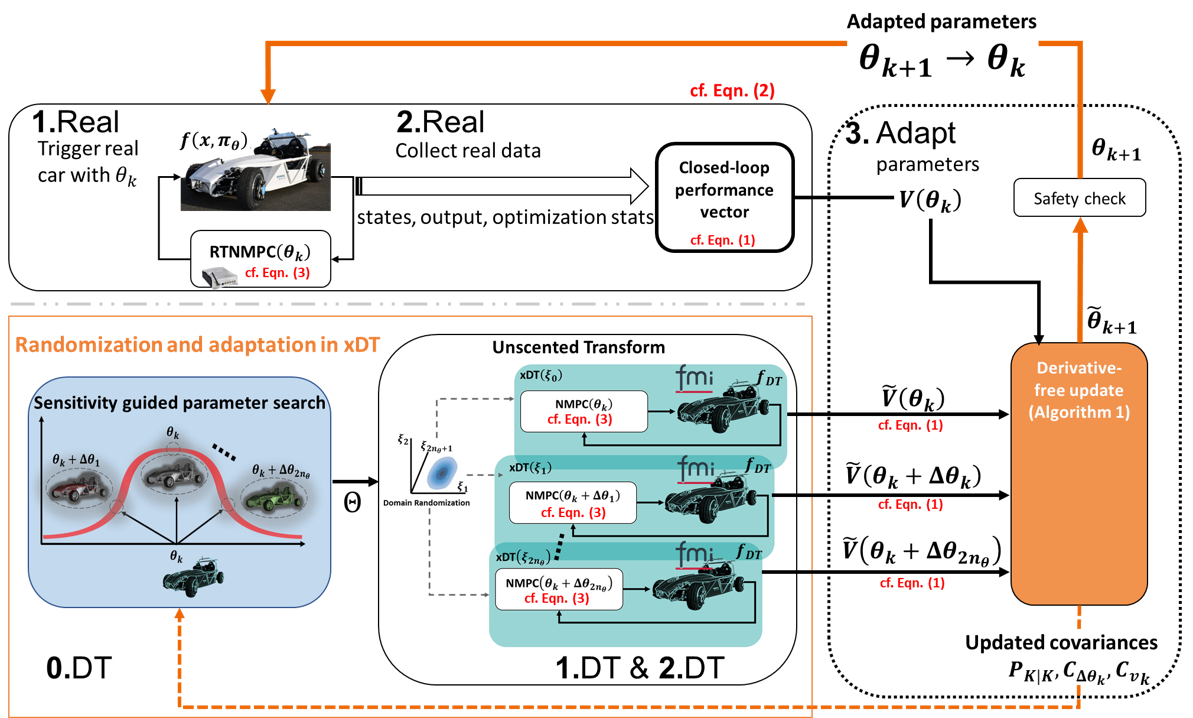}
	\end{adjustbox}
	\caption{\textcolor{responsecolor}{Adaptation framework}: optimize the target domain (real-world) closed-loop performance $V$ safely and iteratively by sampling the different \ac{xDT}s running in parallel with each other and with the real car. Framework combining Domain Randomization, Adaptation and High-fidelity simulation, and driven by real-world performance data.} 
	\label{fig:adaptation_framework}
\end{figure*}

In this work, we propose adapting the parametrizable AD controller by sampling multiple \ac{xDT}s built with Simcenter Amesim that run in parallel as in Figures~\ref{fig:general_adaptation_framework} and~\ref{fig:adaptation_framework}. \textcolor{responsecolor}{The prior knowledge captured by the \ac{xDT}s in terms of the closed-loop controller performance is updated through additional data from real-world feedback, closing by this the Sim2Real gap.}
The \ac{xDT}s are  instantiated in the form of a Functional Mock-up Unit (FMU) for co-simulation, each with its own integration solver. FMUs follow an industry standard to interface and exchange dynamic simulation models. By setting the \ac{xDT} as an FMU, it is possible to create many instances of the plant model each with its own model parameters. \textcolor{responsecolor}{Thus, we are capable of parallelizing over the CPU or GPU, several rollouts of the controller in the loop with an \ac{xDT}, namely a simulated oracle. For this work, we parallelize over the CPU. That is every oracle runs on a standalone core, permitting simultaneous exploration in the parameter space. For a reasonable number of control parameters, the parallelization results in significant CPU time reduction, allowing the real-time application.} We utilize OpenMP, a multithreading implementation in C++ that allows shared-memory multiprocessing~\cite{chandra2001parallel}.

\textcolor{responsecolor}{Unlike the TiL work of~\cite{twin_in_the_loop_23} which focuses on tuning an additional compensator controller such that a certain DT output resembles the real counterpart, we consider the direct performance optimization without explicit quanitifcation of the Sim2Real gap. We address the Sim2Real gap between the DT and the real counterpart by means of DR and by accounting for the uncertainty on the Sim2Real gap within the parameter optimization. Our objective is to 1) extend the approach to multiple TiL for a better control parameter exploration, and 2) minimize the effect of Sim2Real gap by considering multiple \ac{xDT}s with different contextual parameters to robustify the controller against possible real-world realizations. Moreover, the usage of vanilla BO is sample inefficient for a large number of parameters. Often with BO, tuning is either done purely in simulation then transferred to the real-world, or directly carried in the real-world in the form of online learning. That is, the target system is employed for explorative purposes as the parameters vary. We aim to combine a sensitivity guided active exploration on the \ac{xDT}s with the exploitation on the real system, ensuring that the real system lies within the distribution of the simulated performances.}

\textcolor{responsecolor}{By executing a twin of the \ac{NMPC} (cf.~\eqref{eq:lowlevel_bilevelOpt}) with carefully selected control parameters on multiple independent and parallel \ac{xDT}s, we evaluate the distribution of the performance and its sensitivity with respect to the parameters. The performance of the estimated optimal parameters is evaluated in the high-level problem in~\eqref{eq:highlevel_bilevelOpt} with real measurements. This process aims to automatically improve the real-world performance. Experimental results in Section~\ref{section:adaptation_experimental} demonstrate that the framework significantly enhances performance in just few online runs with the real vehicle.}

\section{Learning and adaptation framework}
\label{section:learning_framework}
The goal of the proposed work is to optimize the closed-loop performance in the target domain, while avoiding sampling the different parameter variations directly in this domain, but rather in the \ac{xDT} source domain. The \ac{xDT} environment in this case is composed of both the car model and the AD module to be tuned, the \ac{NMPC}. This simulated oracle mimics the target closed-loop settings, i.e. the real system oracle. The resultant framework should be able to directly deal with external uncertainties $\xi$ that affect the performance, by compensating for their effect without estimating them. On a high level, we seek to minimize the norm of the expected overall performance measure $V$. This is done by estimating the optimal parameters $\theta$, after receiving data feedback from the real oracle measurements over a time window $t_0, \dots, t_0+T$. In this section, we first present the black-box optimization formulation for automatic controller calibration. Further we introduce the approach to push exploration towards \ac{xDT}s and combine it with real-word data for safe and efficient automatic parameter calibration as in Figure~\ref{fig:adaptation_framework}.
\textcolor{responsecolor}{\begin{rem}
We refer to safety as two components. First, the burden of simultaneous parameter exploration is offloaded towards simulation. This allows to explore and detect possibly unsafe regions, and deduce the sensitivity of the performance with respect to the parameters without an online learning phase. Second, as the structure of the underlying \ac{RTNMPC} does not change, we consider safety as the autonomous system is driven by an underlying safe and deterministic controller. This comes in comparison with learning-based controllers that are often based on neural networks for policy generation. Therefore, \ac{NMPC} acts as a safety filter for the adaptation framework.
\end{rem}}

\subsection{Black-box optimization formulation}
The target system, \textcolor{responsecolor}{for which the controller is to be tuned}, evolves according to the unknown dynamics $f$, while its DT evolves according to $f_{DT}$. The map $\mathcal{F}$ defines the system evolution from time $t_0$ to $t_0+T$. The function $\mathcal{H}_i$ is an output transformation map for a specific output performance $Y_i$ out of $n_{O}$ output performance vectors.  Additionally, $W$ and $O$ are the process and output noise. We define a closed-loop oracle:%
\begin{subequations}\label{eq:stochastic_dynamics_equation}%
	\begin{align}
		X_{t_0:t_0+T} &= \mathcal{F}(X_{t_0},U_{t_0:t_0+T}(\theta_k), W_{t_0:t_0+T}),\\
		Y_{i,t_0:t_0+T}  &= \mathcal{H}_i(X_{t_0:t_0+T}, U_{t_0:t_0+T}(\theta_k)) + O_{t_0:t_0+T}, \\
		V(\theta_k) &= \begin{pmatrix}
			Y_{1,t_0:t_0+T} - Y^r_1\\
			\vdots\\
			Y_{n_O,t_0:t_0+T}-Y^r_{n_O}
		\end{pmatrix}.
	\end{align}
\end{subequations}%
\textcolor{responsecolor}{The equations in~\eqref{eq:stochastic_dynamics_equation} define a black-box oracle, or a rollout of the autonomous system with the controller tuned with the parameters $\theta$. The first equation shows the evolution of the internal states $X$ from $t_0$ to $t_0+T$, based on the input trajectory $U(\theta)$ and under the effect of noise $W$. The input $U(\theta)$ results from the parametrizable controller. The second equation calculates the $n_O$ output trajectories $Y_i$, using the measured or monitored signals. The third equation represents the vector $V(\theta_k)$ which stacks the errors of the individual output performances with respect to their desired or reference values $Y^r_i$. We consider the window $[t_0, t_0+T]$ to contain $N_T$ discrete elements such that $N_T=T/T_s$ where $T_s$ is the update period of the measurements and control action. Importantly, in the case of the real system, the true $V(\theta_k)$ is captured as the system evolves with the dynamics $\mathcal{F} = f$ as in Steps 1.Real and 2.Real in Figure~\ref{fig:adaptation_framework}. For the \ac{xDT}s, the approximate performance measure $\tilde{V}_\xi(\theta)$ is calculated and measured similarly to the real system, but with a system evolving based on $\mathcal{F} = f_{DT}$, as in Steps 1.DT and 2.DT of Figure~\ref{fig:adaptation_framework}. The nominal dynamics $f_{DT}$ captured within the DT are highly accurate to evaluate the local behavior of the cost with respect to different parameter combination, but may not perfectly capture the real system dynamics.}

\textcolor{responsecolor}{Furthermore, we consider three level of system dynamics with an increasing level of complexity and accuracy: 1) a simplified and analytical model $\hat{f}$ such as the dynamic bicycle model, used within the low-level \ac{NMPC} controller for motion control and autonomous driving, 2) a multi-physics data-driven high-fidelity simulator $f_{DT}$, namely the \ac{DT} of the car, with 15 degrees-of-freedom used for controller validation in \ac{MiL} and as an \ac{xDT}, and 3) the complex and unknown dynamics on the real vehicle $f$.}

\textcolor{responsecolor}{The inner loop affected by the change of parameters $\theta$ is the parametrizable \ac{AD} controller, in this case the \ac{NMPC} policy $\pi_\theta$ which determines the control action $U$ for a state measurement $X_{t_0}$.
The \ac{NMPC} prediction model evolves according to $\hat{f}$, with a horizon length $T_H$, initial condition $\bar{x}$ and subject to constraints on the inner states $x(t)$ and control $u(t)$ defined by the set $\mathbf{X}$. }
We form the controller parameter calibration problem of Figure~\ref{fig:general_adaptation_framework} as follows:  
\begin{align}%
	\label{eq:highlevel_bilevelOpt}
	\minimize_\theta \E_{\xi}&\ \bigg[\frac{1}{2}\lVert V(\theta)\rVert ^2\bigg] \\ \nonumber
	\stt&\ \theta \in \mathbf{C}_\theta, \\
	x(t+1) = &f\big(x(t), \pi_\theta(x(t)),\xi\big), && t=t_0,\ldots,T\nonumber
\end{align}%
\begin{align}%
	\label{eq:lowlevel_bilevelOpt}
	\pi_\theta(x) = \argmin_{x(.),u(.)}&\bigg[J_{\theta}^*= \int_{0}^{T_H}(x^\top Q_\theta x + u^\top R_\theta u) dt\bigg]\\\nonumber
	&x(0) = \bar{x},\\\nonumber
	&\dot{x} = \hat{f}(x,u)\;\; \forall t\in[0,T_H],\\\nonumber
	& x,u \in \mathbf{X},\nonumber
\end{align}%
where:%
\begin{itemize}%
	\item the vector $V$ is a stacking of performance measures such as passenger comfort, tracking performance, driving style, optimization statistics, \textcolor{responsecolor}{that can only be measured once a closed-loop oracle of the system with the controller. The vector is measured through sensor readings, user feedback or through system performance metrics.}
	\item The vector $\xi$ captures unmodeled dynamics, real-world uncertainties, disturbances, and domain shifts that influence the behavior of the system. The vector $\xi$ can be seen as perturbations to the simulator’s physical parameters (for e.g. mass, inertia, length, friction coefficients), measurement and control noise, and task choice. By considering those perturbations as DR, the trained policy robustifies against real-world uncertainties and avoids overfitting the simulation.  
	\item The vector $\theta$ contains the \ac{NMPC} tuning parameters (in this example the state and input weights $Q_\theta, R_\theta$).
\end{itemize}%

The outer loop~\eqref{eq:highlevel_bilevelOpt} optimizes the target domain performance measured in the real-world as shown in 1.Real and 2.Real of Figure~\ref{fig:adaptation_framework}. \textcolor{responsecolor}{It is possible to measure $V(\theta)$ as a black-box oracle on the real plant, but it is expensive. Instead, we offload the burden towards the \ac{xDT}s where we measure an approximate distribution $\mathbf{\tilde{V}}(\theta)$ as shown in Figure~\ref{fig:general_adaptation_framework}. The algorithm explores and approximates through the DT oracles, the distribution of $V$ and that of the parameters $\theta$ by sampling the \ac{xDT}s in the loop as shown in 1.DT and 2.DT of Figure~\ref{fig:adaptation_framework}.} The inner loop~\eqref{eq:lowlevel_bilevelOpt} of the optimization involves the real-time capable parametrizable NMPC (RTNMPC) that is attached to each \ac{xDT} and to the real car. Each vehicle (real, or \ac{xDT}) is controlled with a clone of the \ac{NMPC} but with different parameters, denoted as $\textrm{NMPC}(\theta)$ that leads to $\pi_\theta$. For clarity, we refer to $\textrm{NMPC}(\theta)$ as the \ac{NMPC} controller parametrized by $\theta$ and that is solved according to~\eqref{eq:lowlevel_bilevelOpt}. The controller to tune or calibrate is available in both its digital and RT versions, as a C++ standalone library. Note that if a different controller architecture is used, for e.g. a Proportional-Integral-Derivative (PID), the inner loop in~\eqref{eq:lowlevel_bilevelOpt} is replaced by a simpler formulation: $\pi_\theta(x) = \theta_Pe(t) + \theta_I\int e(t)dt + \theta_D\dot{e}$, where $e(t)$ is an error with respect to a state or output reference.

\textcolor{responsecolor}{The vector $V$ is computed from onboard measurements such as position tracking errors using a GPS, velocity tracking error using an IMU, comfort level using acceleration and jerk measures, as well as monitored signals such as the computation time of the \ac{NMPC} controller. More precisely, $V(\theta)$ stacks the error of the measurements $Y_i$ with respect to their desired or reference values $Y^r_i$ as in~\eqref{eq:stochastic_dynamics_equation}.
Each $Y_{i}$ is performance indicator of some target output over the closed-loop data collected between $t_0$ and $t_0+T$, representing different metrics to be optimized over. We consider three metrics: $Y_{path}$ collects the path tracking error with $Y^r_{path} = 0$, $Y_{velocity}$ collects the velocity tracking error with $Y^r_{velocity}=0$ and $Y_{cost}$ stores the \ac{NMPC} cost with $Y^r_{cost}=0$. \textcolor{responsecolor}{In practice, $V$ could either contain sparse or dense measurements. For dense measurements: as the closed-loop system evolves, we measure the instantaneous performances (for e.g. tracking error, comfort). For sparse measurements: at the end of a window $T$ a sparse performance metric could also be given, for e.g. the total distance covered during $T$ seconds.  Finally, the total \ac{KPI} combines the individual metrics for multi-objective optimization:
\begin{equation}
	\textrm{KPI} = \frac{1}{2N_T}\lVert V(\theta)\rVert ^2
\end{equation}
}
The \ac{NMPC} controller is implemented in a receding horizon fashion. At every instant, the states are measured (real or virtual), \ac{NMPC} plans a trajectory for a horizon $T_H$ ahead, and the first control action of the trajectory is applied as a policy. As we expose the \ac{NMPC} parameters such as the weighting matrices in the cost function, the choice of $\theta$ directly affects the performance of the controller in the policy $\pi_\theta(x)$. We present in Section~\ref{subsec:gradient_free_algo} an iterative framework to solve problem~\eqref{eq:highlevel_bilevelOpt}, by sampling a carefully chosen set of parameters in the digital world as seen in Steps 0.DT, 1.DT and 2.DT of Figure~\ref{fig:adaptation_framework}. Finally, the derivative free algorithm is presented in Algorithm~\ref{alg:ukf_parameter_estimation}, constituting of the simultaneous exploration in the \ac{xDT}s through a sensitivity guided parameter search as in 0.DT of Figure~\ref{fig:adaptation_framework}, and the performance improvement in the real world by feeding back the real-world data from an oracle with the candidate $\theta_k$ as in 1.Real of Figure~\ref{fig:adaptation_framework}}.
\textcolor{responsecolor}{
\begin{rem}
Although it should be scalable to tune the parameters $\theta$ that affect the constraints $\mathbf{X}(\theta)$ and the prediction model $\hat{f}(x,u,\theta)$ of the \ac{NMPC}, in the remainder of this paper we focus only on automatically calibrating the cost function parameters $Q_\theta, R_\theta$. That is the inner loop in~\eqref{eq:lowlevel_bilevelOpt} is only parametrized by $\theta = [Q_\theta, R_\theta]$, and all other components such as the constraints, the model parameters, or the horizon length $T_H$ are fixed. 
\end{rem}}
\textcolor{responsecolor}{As the \ac{NMPC} is limited in its forecast by the prediction horizon $T_H$, it is rather complicated to assess the effect of a decision taken by the \ac{NMPC} at time $t$, on the performance of the vehicle at time $t+T$ where $T \gg T_H$. For this reason, we seek to solve~\eqref{eq:highlevel_bilevelOpt} where $V$ encapsulates the collected closed-loop data over a sliding time window $T$. The positive scalar $T$ is the window of real-world data collection over which the algorithm tries to improve the performance. It is equivalent to the period of parameter adaptation in the target domain. There exists a trade-off between short and long adaptation period $T$: small values of $T$ lead to local and short term performance improvement over the current driving area with frequent adaptation but with a forgetting factor. On the other hand, large values of $T$ lead to overall performant controller, with less computational demand for the adpatation, but with slower response to local changes. As we do not aim to build a surrogate of the optimization function as with BO, employing the algorithm for adapting the control parameters at a fast rate improves the performance locally but has a forgetting factor when revisiting the same area. Whereas, if long adaptation period is chosen, such as $T=T_{episode}$ where $T_{episode}$ is a full scenario, the algorithm optimizes for one set of parameters that is optimal over the whole scenario.} However, the proposed method allows both by exposing the adaptation rate as a user chosen hyperparameter. \textcolor{responsecolor}{The experiments to be carried in this work are such that we optimize over the control parameters after collecting data for $T = 85$s, allowing the vehicle to launch and engage in path and velocity tracking, as well as completing the parking maneuver. The goal is to optimize over one set of control parameters that works well for a variety of references (i.e. not a repetitive task), and speed profiles.}

\subsection{Research questions}
The stochastic optimization problem to shape $V$ has three hard conditions making it difficult to solve: 
\begin{itemize}
\item \textbf{R.1} the vector $V$ could be of any form, not necessarily differentiable nor analytical, and can only be measured by sampling a rollout.
\item \textbf{R.2} The method must tolerate inexactness in the gradient approximation as $V$ is corrupted with noise.
\item \textbf{R.3} The goal is to optimize directly, but as the adaptation is completed online and the AD module operates on a real-time hardware, sample and data efficiency are a major pillar of the method. It is impossible to sample every trial from both a safety and cost perspective~\cite{frohlich_contextual_2022}. 
\end{itemize}
We seek to solve~\eqref{eq:highlevel_bilevelOpt} iteratively, where the update iteration is denoted by the subscript $k$. Following a target system oracle as in~\eqref{eq:stochastic_dynamics_equation}, $V(\theta_k)$ is measured. As the output is corrupted with noise $O_{t_0:t_0+T}$, let $v_k$ be the stacking of the noise components over the time window and over the individual outputs. The elements of $v_k$ are assumed to be random variables following a Gaussian distribution with zero mean and a prior covariance $C_v$, where $v_k \sim \mathcal{N}(0,\,C_v)$. That is, let $\mathcal{H}(\theta_k)$ be the stacking of all the individual and instantaneous outputs $\mathcal{H}_i$ for the time window $[t_0,t_0+T]$, and $Y^r$ the stacking of the reference values for each of the instantaneous outputs. It follows that the update step is rewritten as:
\begin{align}
	V(\theta_k) &= (\mathcal{H}(\theta_k) - Y^r) + v_k,\\
	\theta_{k+1} &= \theta_k + \Delta\theta_k.
\end{align}
Moreover, the uncertainty on $\theta_k$ results from it following the distribution $\Delta\theta_k \sim \mathcal{N}(0,\,C_{\Delta\theta})$. Note that the distribution of $\theta_k, \Delta\theta_k, V(\theta_k)$ will be approximated through an \ac{UT}, which is not restricted to the assumption that the distributions of noise sources are Gaussian~\cite{julier_ukf}.

\subsection{Sim2Real using xDT and derivative-free optimization}
\label{subsec:gradient_free_algo}
The proposed algorithm falls under the category of directional direct search methods that sample at several perturbed points to explore the local behavior of the cost, then form a directional step.
However, to avoid greedy steps by choosing the sample point with the minimum function evaluation, we consider the distribution over all $\theta$ and $\xi$ and step in the direction of expected value improvement.   
For this, we adopt a zeroth order optimization framework where we only have access to the value function $V$. Moreover, scalability to both episodic and rapid local adaptation is sought for by the method, which is not possible by BO. As we seek to extend the training to any desired measurable objective, differentiable or not, and to exploit parallel \ac{xDT} black-box structure, we consider gradient or Derivative Free, zeroth order Optimization (DFO). A list of available methods for such an approach is presented in~\cite{larson_menickelly_wild_2019}. One method that falls in this category is the optimization through gradual parameter estimation in the form of iterative Kalman filtering. In fact, \ac{EKF} is proven to work for such applications as asymptotically it resembles a gradient method with diminishing step-size. From~\cite{Bertsekas/99}, \ac{EKF} resembles a Gauss-Newton step for least squares problems, except that it works incrementally, and for large iterations the method leads to a Gauss-Newton step with diminishing step size\cite{Bertsekas/99}.

In our case we propose the use of iterating \ac{UKF} steps, with no linearization involved, to deal with arbitrary complex $V$ and incorporate real-world data.
In the work of~\cite{9011159}, the \ac{IUKF} is presented as a derivative free extension to the \ac{IEKF}, requiring no linearization and that can be used in optimization algorithms. As the gradient is never numerically calculated, the \ac{UT} prediction and estimation steps allow to form a sort of zero-order derivative-free update step. \textcolor{responsecolor}{However, further studies are to be carried to extend the results of resemblance between \ac{IEKF} and a Gauss-Newton algorithm, towards the \ac{UKF} and its ability to accurately solve a nonlinear least squares type of optimization}. 

Exploiting the UKF for optimal parameter tuning has been presented in~\cite{menner2023automated} to outperform BO given its ability to handle high-level of noise through the weighted averaging of samples. Moreover, the number of samples in a UKF framework grows linearly with the number of parameters and is model free as it does not require to build a surrogate of the performance metric. Finally, the optimal parameter search is guided through the covariance matrices eliminating the need for dangerous random sampling. In~\cite{ALLAMAA2022385}, the combination of two gradient-free methods (UKF and SPSA) were presented to estimate the optimal set of control parameters $\theta$ in the target domain. The derivative-free update is achieved through sampling several \ac{xDT}s online by smartly perturbing the parameters around the current iterate $\theta_k$, and measuring the performance index $\tilde{V}_\xi(\theta+\Delta\theta)$ in each \ac{xDT}. We extend~\cite{ALLAMAA2022385} by providing an experimental validation of the approach through a parallelizable framework, to automatically adapt the parameters of a RTNMPC
controller used for autonomous valet parking applications. Moreover, we complete the algorithm by adding an adaptation layer to the covariance matrices that are directly related the learning rate of the approach, and its convergence. For robustness and generalization of the controller against external uncertainties $\xi$, we add DR in the \ac{xDT} rollouts in the form of noise and model parameter uncertainties. For online applications, both UKF and SPSA are suited for parallel computation as the sample rollouts in the \ac{xDT}s are independent. The iterative DFO algorithm is presented in Algorithm~\ref{alg:ukf_parameter_estimation}.
\subsubsection{\textbf{Unscented Kalman Filter}}%
An UKF propagates the parameters through nonlinear dynamics and updates the estimated parameters without relying on gradient measurements. The method samples a set of points around the current estimate and can be used for estimating the optimal controller parameters as in~\cite{menner2023automated}. The UKF is chosen over an EKF, as the former samples several points simultaneously. In an iterative scheme, UKF forms a more accurate estimate of the stochastic directional derivative between the parameters to tune and the performance index to optimize over. \textcolor{responsecolor}{This is possible as UKF captures the uncertainty by using a set of sigma points, instead of an explicit linearization around the mean estimate. This generates a set of simulation models instead of a single one, allowing it to better capture nonlinear systems and a wider range of distributions~\cite{julier_ukf}}. In the presence of nonlinearities and inaccuracies, UKF tends to be more robust than EKF as it avoids linearization around the mean estimate of the Gaussian and can better propagate the uncertainties. \textcolor{responsecolor}{Moreover, at the cost of computational overhead, UKF provides generally better accuracy in parameter estimation. It utilizes not only the mean estimate for a rollout, but rather a set of carefully chosen sigma points to capture the mean and covariance of the distribution}. This makes it attractive for frameworks combining black-box plant models and possibly non-analytical, nonlinear evaluation metrics.
In~\cite{ALLAMAA2022385}, we proposed to propagate the \ac{UT} through the \ac{DT} as the predicted performance is closer to the target domain than simplistic models and allows the algorithm to converge faster. 

Given an $n_{\theta}$-sized parameter vector $\theta$ following a normal Gaussian distribution of the form $\theta \sim \mathcal{N}(\theta_k,\,P_{k|k})$  at an instance $k$, we obtain a set of candidate sampling points $\Theta$ \textcolor{responsecolor}{around the mean $\theta_k$. The sampling or sigma points are concatenated in a matrix $\Theta =[ \theta_k  \vert \theta_k + c_kA^{j}  \vert \theta_k - c_kA^{j}]\in \mathbb{R}^{n_\theta\times 2n_\theta+1}$}, where $c_k = c_0 =  \sqrt{n_\theta + \lambda}$, and $A^j$ is the $j^{th}$ column of the matrix $A = \sqrt{P_{k|k}}$. The matrix $A$ is computed by performing a Cholesky decomposition of the prior covariance matrix $P_{k|K}$.
The scalar $\lambda$ dictates the spread of the sampling points around the mean and contributes to fine-tuning the higher order moments of approximation~\cite{julier_ukf}. From~\cite{julier_ukf}, one good heuristic for choosing $\lambda$ is $n_\theta + \lambda = 3$ \textcolor{responsecolor}{which minimizes the difference between the moments of the standard Gaussian and the sigma points up the fourth order.} \textcolor{responsecolor}{As from $\Theta$, the sampling points are generated from the step size $c_k$ in the direction of the uncertainty represented by $A^j$}.  In total, $2n_\theta+1$ rollouts are performed for one update step $k$ to $k+1$. As it is inefficient to evaluate every sample point $\Theta^j$ in the target domain, we propagate the dynamics and performance evaluation through the \ac{xDT}s. Then, we form the distribution in the form of mean and covariance as a weighted sum of the independent and parallel \ac{xDT} closed-loop performances, through the weighting parameters $w_a$. The weighting vector is formed as $\mathcal{W} = [w_a^0,\, w_a^1,\, \dots,\, w_a^{2n_\theta}] \in \mathbb{R}^{1\times 2n_\theta+1}$, where:%
\begin{align}\label{eq:ukf_weighting_pts}%
	w_a^0 = \frac{\lambda}{(n_\theta + \lambda)}, \,
	w_a^j = \frac{1}{2(n_\theta+\lambda)} \quad \textrm{for } j=1,\dots,2n_\theta.
\end{align}
The scalar $w_a^0$ is the weight associated with the first sample point, i.e. the mean $\theta_k$, and $w^j_a$ is the weight associated with the $j^{th}$ sigma point. 
As the closed-loop dynamics in the target domain evolve according to $f(x,\pi_\theta)$ from time $t_0$ to $t_0+T$, all the \ac{xDT}s are spawned and run in parallel with it starting from the initial condition $x(t_0)$. At time $t_0+T$, the algorithm calculates the stochastic directional derivative in the form of a Kalman gain $K_k$ which contains first and second order information about the step direction for performance improvement. Furthermore, the gain $K_k$ incorporates the covariance of the parameters, and cross-covariance between the parameters and the predicted output. \textcolor{responsecolor}{Drawing resemblance to the equivalence between IEKF and Gauss-Newton, the spread of the covariance matrix can be seen as the step size in the Gauss-Newton algorithm}. Finally, as we aim for a bidirectional communication between real and digital, the target domain performance data is received in the adaptation framework. \textcolor{responsecolor}{The parameters are updated by combining the Kalman gain $K_k$ from the \ac{xDT}s exploration with the target performance $V(\theta_k)$}, to generate the next possible iterate $\tilde{\theta}_{k+1}$:%
\begin{equation}
	\tilde{\theta}_{k+1} = \theta_k - K_kV(\theta_k).
	\label{eq:kalman_update_step}
\end{equation}%
\textcolor{responsecolor}{Note that $\tilde{\theta}_{k+1}$ undergoes a safety check to be discussed in~\ref{subsec:safety_check} before being applied as $\theta_{k+1}$.}

\begin{algorithm}[t]
	\caption{Stochastic DFO algorithm}\label{alg:ukf_parameter_estimation}
	\begin{algorithmic}
		\Require $\Theta, w_a,  C_{\Delta\theta,0}, C_{v,0}, P_{0|0}$
		\State \hskip-1em \textbf{Step 1: Unscented Transform}
		\State $\bar{\theta} = \theta_k = \sum_{j=0}^{2n_\theta}w_a^j\Theta^j$
		\State 	$P_{k+1|k} = C_{\Delta\theta,k} + \sum_{j=0}^{2p}w_a^j(\Theta^j - \bar{\theta})(\Theta^j - \bar{\theta})^T$
		\For{$j = 1,\dots,2n_\theta$}\Comment{Parallel xDT rollouts}
		\State $y^j =  \tilde{V}_{\xi}(\Theta^j)$\Comment{Propagate}
		\EndFor
		\State $\bar{y} = \sum_{j=0}^{2n_\theta}w_a^jy^j$

		\State \hskip-1em \textbf{Step 2: Cross covariance update}
		\State $P_{\theta y} = \sum_{j=0}^{2n_\theta}w_a^j(\Theta^j - \bar{\theta})(y^j - \bar{y})^T$
		\State $C_{yy,k} = \sum_{j=0}^{2n_\theta}w_a^j(y^j - \bar{y})(y^j - \bar{y})^\top$
		\State $P_{yy} = C_{v,k} + C_{yy,k}$

		\State \hskip-1em \textbf{Step 3: Calculate gain and curvature}
		\State $K_k = P_{\theta y}P_{yy}^{-1}$ \Comment{Kalman gain}
		\State $P_{k+1|k+1} = P_{k+1|k} - K_kP_{yy}K_k^\top$ \Comment{Posterior covariance}

		\State \hskip-1em \textbf{Step 4: Calculate the optimization step}
		\State $\delta\theta_{SPSA} = -a_k \hat{g}_{SPSA} $ from~\eqref{eq:spsa_gradient} and ~\eqref{eq:update_param_SPSA} \Comment{SPSA step}
		\State $\delta\theta_{UKF}  = -K_kV(\theta_k) $ \Comment{UKF step}
		\State $\Delta\theta_{k}  = w\delta\theta_{UKF} + (1-w)\delta\theta_{SPSA} $ \Comment{UKF+SPSA}
		\State $\tilde{\theta}_{k+1} \leftarrow \theta_k +\Delta\theta_{k} $
		\State $a_k \leftarrow a/(\lVert y^{j=0} \rVert^2 + k^{0.602})$
		
		\State \hskip-1em \textbf{Step 5: Adapt noise covariances from~\cite{8273755}}
		\State $\epsilon_k = V(\theta_k) - \bar{y}$
		\State $C_{\Delta\theta,k+1} \leftarrow \alpha C_{\Delta\theta,k} + (1 - \alpha)(\Delta\theta_k \Delta\theta_k^\top)/(k^2)$
		\State $C_{v, k+1} \leftarrow \alpha C_{v,k} + (1-\alpha)(C_{yy,k} + \epsilon_k^\top\epsilon_k)/(k^2)$ 
		
	\end{algorithmic}
\end{algorithm}

\subsubsection{\textbf{Simultaneous Perturbation Stochastic Approximation (SPSA)}}%
There exist several approaches for estimating the stochastic gradient of a certain performance index with respect to $n_\theta$ parameters, out of which the SPSA method that was first proposed in~\cite{Spall1998ANOO} for stochastic optimization. By perturbing all parameters simultaneously, SPSA requires exactly only 2 evaluations of $L(\theta_k) = \lVert V(\theta_k) \rVert^2$ for the stochastic approximation of the gradient, regardless of $n_\theta$. The scalar $L(\theta_k)$ represents the total key performance index to minimize in~\eqref{eq:highlevel_bilevelOpt}. The method has proven to be beneficial in case no analytical relationship between $V$ and $\theta$ is present and for online deployment purposes, as it is memory and data efficient.  Moreover, as sampling $V$ with the perturbed $\theta$ directly in the target (real world) is unsafe and non-viable, we perform a simulation based optimization  to solve for $\delta L/\delta\theta_k=0$ by estimating the gradient $g(\theta) = \nabla L(\theta)$ while simultaneously perturbing all parameters: 
\begin{equation}\label{eq:spsa_gradient}
	\hat{g}_{SPSA}(\theta_k) = \frac{L(\theta_k + c_k\Delta_k) - L(\theta_k - c_k\Delta_k)}{2c_k}
	\begin{pmatrix}
		\Delta_{k1}^{-1}\\
		\vdots\\
		\Delta_{kn_\theta}^{-1}
	\end{pmatrix}
\end{equation}
\textcolor{responsecolor}{The gradient estimate in~\eqref{eq:spsa_gradient} differs from a usual finite difference approximation in that it employs only 2 perturbed measurements instead of $2n_\theta$.}
Every parameter is independently perturbed with a magnitude of $c_k\Delta_{kj}$ where $c_k$ is the differential step size hyperparameter of the SPSA algorithm. \cite{Spall1998ANOO} suggests a random perturbation vector $\Delta_k = [\Delta_{k1},\dots,\Delta_{kn\theta}]^T$, following a Bernoulli distribution symmetric about zero, with mutually independent elements. 
The recursive SPSA algorithm updates the parameter estimate, using a step size $a_k$ as:%
\begin{equation}\label{eq:update_param_SPSA}%
	\tilde{\theta}_{k+1} =  \theta_k - a_k\hat{g}_{SPSA}(\theta_k)
\end{equation}%
From~\cite{Spall1998ANOO}, a necessary condition for the convergence of the algorithm is that both the update and sampling steps ($a_k$, $c_k\Delta_k$) converge to zero. In other words, $a_k, c_k > 0 \; \forall \; k$, and $a_k \rightarrow 0, c_k \rightarrow 0 \; \textrm{as} \; k\rightarrow \infty$.  Therefore, we propose to combine UKF and SPSA, by setting the SPSA sampling to be guided by the real data-driven covariance of the parameters $P_{k|K}$ from the UKF counterpart:
\begin{equation}
	c_k \Delta_k = (c_0 \times \sqrt{P_{k|k}}) \times {\rm Bernoulli}(1), 
\end{equation}
where $\Delta_{k}$ uses a Bernoulli $\pm1$ distribution. 
The parameter search will then be fully guided by the parameters' uncertainty, to generate sampling points for SPSA. Starting from an initial scalar $c_0$, we can now ensure a diminishing sampling step condition, if the uncertainty diminishes as will be explained in the next section. 
Moreover, the update step size $a_k$ is set to diminish as per Step 5 of Algorithm~\ref{alg:ukf_parameter_estimation}. The division by $y^{j=0}$ scales the step size, and ensures a continuous parameter adaptation, if the performance index is non-zero. The factor $k^{0.602}$ follows the guidelines in~\cite{705889}.  
Note that since SPSA does not require calculating gradients analytically, only relies on zero-order information of the cost to optimize $V(\theta_k)$, and can handle noisy and complex objective functions, it is a derivative-free method and is suitable to address our research questions \textbf{R.1}, \textbf{R.2}.
\textcolor{responsecolor}{
\subsubsection{\textbf{Fused adaptive UKF with SPSA}}%
An extensive convergence theory for SPSA exists and pertains to both local and global optimization as in~\cite{spall92_spsa, SPSA_global}. SPSA has shown to be an effective, fast and efficient method for derivative-free global optimization methods and has been extensively used for numerous applications as in~\cite{spsa_applications}. However, SPSA can also be sensitive to the noise realizations over the random perturbation of the parameters. Furthermore, vanilla SPSA does not follow a sensitivity guided parameter search. For these reasons, we propose to combine UKF and SPSA. Given that the UKF algorithm generates the $2n_\theta$ rollouts, we exploit the measurements to perform an aggregate SPSA step. Moreover, as we aim to transfer the performance from Sim2Real, SPSA does not embed by default the characteristic of mixing simulation and real data. On the other hand, the iterative UKF with the adaptive noise covariance captures the Sim2Real gap. It introduces the gap as a bias in the iterative framework, and smartly explores interesting parameter regions, through the sensitivity guided parameter search. Thus, we can exploit the parameter search and rollouts generated by the iterative UKF to generate more accurate SPSA stochastic gradients. SPSA improves the UKF performance. UKF tends to get stuck in local minima because it captures the noise and samples several sigma points around the current mean. The efficient use of SPSA for global optimization as in~\cite{SPSA_global} allows to escape those minima by injecting noise on the gradient. Therefore, we combine the strengths from SPSA and UKF in the parameter calibration: UKF allows generating the sample points and rollouts, and SPSA allows escaping local minima in the optimization routine. Moreover, we fuse the adaptation steps on the parameter estimate from UKF and SPSA (c.f.~\eqref{eq:kalman_update_step},~\eqref{eq:update_param_SPSA}) in Step 4 of Algorithm~\ref{alg:ukf_parameter_estimation} through a weighted mean using the hyperparameter $0\leq w\leq 1$. For the remainder of the work, we set $w=0.5$. However, this hyperparameter can be tuned more carefully if more weight is to be assigned to the step using the current real-world performance with UKF, or the expected simulation improvement from SPSA.}
\textcolor{responsecolor}{
\begin{rem}	
Although the parameter search resembles a Monte Carlo method, the sampling points are not random, but are rather carefully chosen to capture information about the distribution in the exploration~\cite{julier_ukf}. This results in the sampling efficiency when running the multiple simulations, and addresses \textbf{R.3}.
That is, the parallel \ac{xDT}s rollouts follow the propagate step of Algorithm~\ref{alg:ukf_parameter_estimation}. We augment the algorithm with DR in the form of $\xi$ to robustify the controller against external uncertainties when transferring to the real world. In its simplest form, $\xi$ is an additive input and output noise, and some mismatch on model parameters such as the total vehicle mass, the cornering stiffnesses, etc. The \ac{DT}s are sufficient to capture the vehicle behavior under the varying uncertainties, but more importantly serve to capture the sensitivity between the output and the parameters. 
\end{rem}} 

\subsection{Adaptive covariance matrix}
As the data-driven UKF learns the optimal parameter set in a recursive approach, research has proven that the learning rate is directly correlated to the choice of process and output noise covariances $C_{\Delta\theta}, C_v$~\cite{8273755}. Moreover, if those two matrices are kept constant, the parameters' covariance matrix $P_{k|k}$ does not necessarily diminish. This leads to unnecessary sampling in the parameter space. Moreover, as we combine UKF with SPSA, it is imperative that both the sampling and update steps $(a_k$, $c_k\Delta_k)$ diminish as the performance stops improving. For this, we follow the adaptive noise covariance matrices  approach developed in~\cite{8273755} for an EKF in a state estimation framework and introduce it to our UKF approach as seen in Step 5 of Algorithm~\ref{alg:ukf_parameter_estimation}. It consists of adapting the parameter step and output covariances $C_{\Delta\Theta}, C_v$ through a forgetting factor $\alpha$. We set $\alpha=3$. Furthermore, the Sim2Real gap is captured through $\epsilon_k$. \textcolor{responsecolor}{It allows the algorithm to account for the Sim2Real gap in the form of a bias with respect to the simulated distribution of oracles. By adapting $C_v$ on-the-go, we integrate the uncertainty on the Sim2Real gap into the update step. Therefore, the objective is not to develop a DT that mimics the real counterpart, but rather have the real-world closed-loop performance contained within the distribution of digital closed-loop performances. Through the adaptive covariance matrices, the algorithm leads to an uncertainty minimization as $C_{\Delta\theta}, C_v$ and $P_{k|k}$ diminish. This uncertainty minimization contributes to the convergence of the algorithm as the search space shrinks. Eventually, the real-world performance is mimicked by its digital counterpart, hence achieving a Sim2Real transfer. Thus, the adaptive covariances lead to a converging algorithm with shrinking update and exploration step sizes, which are necessary conditions for the convergence of SPSA.}

Combining all the components together, the overall framework of $\DCAUKS$ is shown in Figure~\ref{fig:general_adaptation_framework}, and the particular example of \ac{NMPC} adaptation in Figure~\ref{fig:adaptation_framework}. The underlying iterative and data-driven calibration scheme is presented Algorithm~\ref{alg:ukf_parameter_estimation}. The black-box optimization method is not one shot, but rather iterative: first, data is collected from the target domain. Second the local properties of the optimization function (nonlinearities, curvature) are discovered or learned in the digital domain. Third, the parameter iterate is applied on the target system. Finally fourth, the exploration space is adapted through a feedback on the real output and a sensitivity analysis. The update step is according to Step 4 of Algorithm~\ref{alg:ukf_parameter_estimation}. A summary of the algorithms hyperparameters is found in Table~\ref{tab:parameters_table1}.%
\textcolor{responsecolor}{
\subsection{Safety check}
\label{subsec:safety_check}
As we deal with noisy data, performance improvement is sought over a global optimization. For this, we incorporate a simple safety check on the updated parameters. The safety check consists of rolling out an additional oracle with the nominal DT and the controller with the new parameter set $\tilde{\theta}_{k+1}$. This digital oracle aims to verify that 1) the closed-loop system is not unstable, 2) the parameter set is within its bounds $C_\theta$, 3) the \ac{NMPC} cost function or energy measure is not increasing in comparison with the rollout with $\theta_k$. 
For applications with highly nonlinear dynamics and in
presence of noise, strict Lyapunov stability is hard
to prove. Therefore, the third check consists of checking that the total closed-loop \ac{NMPC} energy or cost function $H^\theta_{cost} = \sqrt{\frac{1}{N_T}\sum_{i=1}^{N_T}(J^*_{\theta}(i))^2}$ is contained. The instantaneous optimal cost $J^*(\theta)$ is calculated as in~\eqref{eq:lowlevel_bilevelOpt}, and $N_T$ is the number of discrete measurements between in $[t_0,t_0+T]$. For this, we exploit the \ac{xDT}s to check that starting from the same initial conditions $\bar{x}$ at $t_0$, an \ac{xDT} rollout with $\tilde{\theta}_{k+1} > 0$ would have kept the total energy contained as $H_{cost}(\tilde{\theta}_{k+1}) \leq (1+R)H_{cost}(\theta_k)$, where $R\geq0$. 
Eventually, $\theta_{k+1} \leftarrow \textrm{Check}(\tilde{\theta}_{k+1})$, otherwise $\theta_{k+1}=\theta_k$. Note that in all the results presented below, $\tilde{\theta}_{k+1}$ were safe enough to be directly exploited on the real system.}
\textcolor{responsecolor}{
We exploit the prior parameter covariance $P_{k|k}$ to modify the step size $c_k$ as in~\cite{ALLAMAA2022385}. When computing $\Theta$, the scalar $c_k$ is adjusted such that the parameters $\theta$ are within the safe bounding box $C_\theta = [\theta_{\textrm{min}}, \theta_{\textrm{max}}]$ provided by the user. That is, we add a regularilization heuristic to choose the smallest $c_k$ such that none of the parameters violates their limits. This is important in the case of \ac{NMPC} where $Q_\theta \succeq 0, \, R_\theta\succ0$, to avoid sampling the \ac{xDT}s with indefinite matrices.  The regularization method avoids
clipping the parameters individually to their limits which breaks the Gaussian distribution. Instead, we find the suitable exploration step size $c_k$ that forms the sigma points $\Theta = \theta_k \pm c_k\sqrt{P_{k|k}}$ such that $c_k = \min(c_k, c|\theta_k \pm c\sqrt{P_{k|k}} \in C_\theta)$. This heuristic ensures that the Gaussian distribution of $\Delta\theta_k$ around $0$ is preserved.}
\section{NMPC adaptation in simulation}
The objective of the data-driven calibration framework is not to replace online learning with offline learning, nor the opposite. We seek to exploit heavily the digital world to generate a prior on the distribution of the parameters and their uncertainties. Then, we complement with the real-world data as to close the Sim2Real gap with minimal real-world effort. We first validate $\DCAUKS$ for an autonomous valet parking application in simulation, where NMPC acts as the low-level vehicle motion controller.
\label{section:adaptation_simulation}
\subsection{Autonomous valet parking formulation}
The underlying vehicle model used within the \ac{NMPC} formulation should be able to capture well the dynamics of the vehicle, without over complicating it such that it remains real-time feasible. For this, we employ the fused kinematic and dynamic single-track model in the Curvilinear (Frenet) frame as presented in~\cite{Allamaa2022SafetyEF}. This frame transformation from a Cartesian to an error frame with respect to the path, allows us to be domain independent and to avoid carrying reference trajectories within the \ac{NMPC}.%
 The vehicle dynamics are over the longitudinal and lateral velocities ($v_x, v_y$) and the yaw rate $r$. The state $s$ tracks the evolution along the centerline, and $w$ and $\theta$ track the distance and heading deviation from the path centerline.
The formulation is parametrized by the curvature $\kappa_c(s)$ which, alongside initial condition, is enough to capture the evolution of the deviations. Moreover, the Curvilinear formulation allows us to convexify position constraints of the vehicle on curved roads as they can be directly cast as a tube constraint of the form $w_l\leq w(t) \leq w_r$. Here, $w_l,\,w_r$ are the left and right limit deviations from the centerline. We augment the curvilinear single-track model with the throttle ($\dot{t_r}$) and steering rate ($\dot{\delta}$) as control input in the \ac{OCP} for smoother driving. Therefore, the state vector in the \ac{NMPC} is $x = [v_x, v_y, r, s, w, \theta, \delta, t_r]$ and the input vector is $u = [\dot{\delta}, \dot{t}_r]$. The dynamics are represented by $\dot{x} = (\lambda)f_{dyn}(x,u) + (1-\lambda)f_{kyn}(x,u)$. We fuse the kinematic $f_{kyn}$ and dynamic $f_{dyn}$ single-track models to allow smooth and differentiable transition from low to high speed and to permit reaching a zero velocity. This is possible through the smooth activation value $\lambda(v_x)$, generating a continuous switching between the models inside the optimization routine. For an autonomous valet parking application, the \ac{NMPC} optimizes as in~\eqref{eq:lowlevel_bilevelOpt} over  $J_{\theta}^* = \min \int_{t_0}^{t_0+T_H} (x-x^r)^\top Q_\theta (x-x^r) + u^\top R_\theta u$, where the reference $x^r = [v^r, \mathbf{0}]$. That is, all states (except for $s$) and control inputs are regulated to zero, except for the velocity that tracks the reference $v^r$. Moreover, the path evolution $s$ is not penalized in the \ac{NMPC} cost (i.e. $Q_s=0$). Therefore, $Q_\theta \in \mathbb{R}^{7\times7}, \, R_\theta \in \mathbb{R}^{2\times2}$. The \ac{NMPC} policy $\pi_\theta(x_{t_0})$ is parametrized by $\theta$ to tune, and where the first control action is applied in a receding horizon scheme. As multiple and single shooting methods were non-viable for our real-time \ac{NMPC} application, we resort to the \ac{SOCSE} method in~\cite{Allamaa2022SafetyEF} to cast the \ac{OCP} into an \ac{NLP} problem. Only box constraints are imposed on all optimization variables $x,u$. A horizon $T_H=3$ s, and splines of order 5 are used within \ac{SOCSE}. 
\begin{rem}
	The detailed \ac{NMPC} formulation has been omitted, as the goal is to automatically calibrate a black-box controller, in a black-box oracle, without particular and detailed knowledge on the controller and its parameters. A detailed \ac{NMPC} formulation is found in~\cite{Allamaa2022SafetyEF}.
\end{rem}

\subsection{Parameter tuning with domain randomization}\label{subsection:parameter_tuning}
The objective is to shape the optimal \ac{NMPC} policy such that it compensates for the finite horizon assumption, by feeding back the closed-loop performance over a larger time window $T > T_H$. We optimize over the vector $\theta$ that stacks the elements of the diagonal matrices $Q_\theta, R_\theta$ such that $\theta = [\text{diag}(Q_\theta), \text{diag}(R_\theta)]$. The optimization objective is to improve the real-world performance that has been deteriorated by the Sim2Real transfer. The main Sim2Real uncertainties are in the low-level actuation system such as steering delays, simplified throttle and brake models, and the road grade of approximately $4\%$ that was never accounted for in the \ac{NMPC} model. \textcolor{responsecolor}{For this, we set $V$ in~\eqref{eq:stochastic_dynamics_equation} to be composed of the measurements over 3 outputs $Y_{velocity}$, $Y_{path}$ and $Y_{cost}$:
\begin{subequations}\label{eq:individual_outputs}
	\begin{align}
		Y_{velocity} &= [v_x(1) - v^r(1),\dots,v_x(N_T) - v^r(N_T)],\\
		Y_{path} &= [w(1),\dots,w(N_T)],\\
		Y_{cost} &= [J^*_{\theta}(1),\dots,J^*_{\theta}(N_T)].
	\end{align}
\end{subequations}}
The objective is to simultaneously improve velocity and path tracking performances, while keeping the energy as low as possible through the inclusion of the \ac{NMPC} cost in the objective to minimize. Each metric $H^\theta$ is a \ac{RMS} over the closed-loop data collected during $[t_0, t_0+T]$ through an oracle with parameter $\theta$. Considering the $N_T$ discrete elements in the rolling window, we calculate the RMS of the individual outputs:
\begin{subequations}\label{eq:RMS_outputs}
	\begin{align}
		H^\theta_{velocity} &= \sqrt{\frac{1}{N_T}\sum_{i=1}^{N_T}(v_x(i) - v^r(i))^2}\\
		H^\theta_{path} &= \sqrt{\frac{1}{N_T}\sum_{i=1}^{N_T}(w(i))^2}\\
		H^\theta_{cost} &= \sqrt{\frac{1}{N_T}\sum_{i=1}^{N_T}(J^*_{\theta}(i))^2}
	\end{align}
\end{subequations}
Importantly, the transformation to the Curvilinear formulation allows to directly penalize the lateral path tracking error $w$. It measures the instantaneous deviation from the path, regardless of the global reference system. This transformation serves as a latent invariant space in a \ac{DA}.
 \tabcolsep=1.5pt\relax
\begin{table}[!t]
	\caption{Summary of hyperparameters for Algorithm~\ref{alg:ukf_parameter_estimation} \label{tab:parameters_table1}}
	\centering
	\begin{tabular}{|c||c||c|}
		\hline
		Parameter & Description & Initial Value\\
		\hline
		$n_\theta$ & Number of parameters to tune & 9\\
		\hline
		$n_O$ & Number of user-defined performance metrics & 3\\
		\hline
		$C_{\Delta\theta}$ & Parameter perturbation covariance  & $\mathbf{I}$\\
		\hline
		$C_{v}$ & Noise covariance & $\mathbf{I}$\\
		\hline
		$P_{0|0}$ & Parameter covariance & $\mathbf{I}$\\
		\hline
		$w_a^0$ & Weight on first sample & -2.0\\
		\hline
		$dt$ & Sample time of measurement [in seconds] & 0.05\\
		\hline
		$N_T$ & Number of steps in adaptation window & 1700\\
		\hline
		$T$ & Adaptation window [in seconds] & 85\\
		\hline
	\end{tabular}
\end{table}

\begin{figure}%
	\begin{center}
		\input{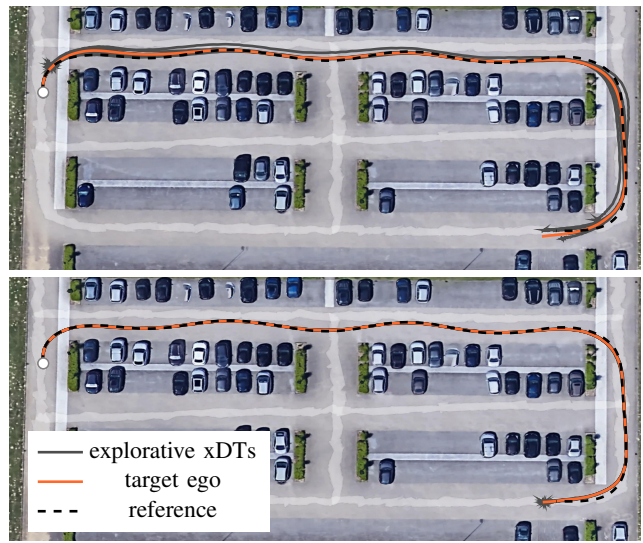}
		\caption{MiL training on dynamic path: Top: Iteration 0: the ego vehicle follows the path with sub-optimal performance. The \ac{xDT}s have a high variance in their performance;
			Bottom: Iteration 1: after one single training iteration, the ego performance has improved and the \ac{xDT}s are overlapped} 
		\label{fig:mil_training_step0}
	\end{center}
\vspace{-0.0cm}
\end{figure}

We first run the calibration process in a \ac{MiL} approach, where the target vehicle is an \ac{xDT} with different parameters from the ones used for sampling. This is to showcase the adaptation from one model to the other (Sim2Sim). We train on one single dynamic path that has been recorded by a human driver as shown in Figure~\ref{fig:mil_training_step0} in dashed-black line. The calibration is started by setting $Q_\theta$ and $R_\theta$ as unit matrices $\mathbf{I}$, that is $\theta=\mathbf{1}\in\mathbb{R}^{9\times1}$. The choice of a unity vector is an uneducated initial guess for the control parameters. It leads to a stable but under-performant oracle. 19 \ac{xDT}s are parallelized and run independently each with a perturbed $\theta_0 + \Delta\theta$. From Figure~\ref{fig:mil_training_step0} on top, the exploration process of the \ac{xDT}s is clear, and they perform with a significant variance as also depicted in Table~\ref{tab:mil_tuning_performance}. Some of the \ac{xDT}s are not able to complete the scenario. However, after just one tuning iteration with a $\DCAUKS$ update and with feedback from the target vehicle performance, the next iterate $\theta_{1}$ is applied. The tracking performance is immediately improved as the RMS on path tracking drops from $0.589\,\textrm{m}$ to $0.206\,\textrm{m}$. The RMS on velocity error, \ac{NMPC} cost as well as the total \ac{KPI} are also decreased. The total \ac{KPI} in~\eqref{eq:highlevel_bilevelOpt} improves by 16\% within one iteration and 70\% within 4 iterations. Moreover, the sampled \ac{xDT}s within only the first iteration have converged to a quasi-similar performance level. All the \ac{xDT}s are superimposed, due to a shrinking uncertainty on the parameter set. Moreover, the performance of the target vehicle is contained within the distribution of the \ac{xDT}s, within one standard deviation. The results of the iterations are indicated in Table~\ref{tab:mil_tuning_performance}. The uncertainty on the tracking performance output drops from $0.296\,\textrm{m}$ to $0.015\,\textrm{m}$ within one iteration. 
From the first iteration, the identified sensitivity between the controller parameters and the performance is not only simulation (source) based, but rather it is enriched with the target system data. The uncertainty on the parameter set shrank directly, and the method sampled more efficiently than with trial and error. 
Given that all the \ac{xDT}s run in parallel with the target vehicle as we presented using an FMU, the adaptation took place in real-time. %

\tabcolsep=1.5pt\relax
\begin{table}[!t]
	\caption{\textcolor{responsecolor}{Performance evolution through the automatic tuning iterations in Figure~\ref{fig:mil_training_step0}: 70\% \ac{KPI} improvement in 4 iterations and minimization of the performance uncertainty\label{tab:mil_tuning_performance}}}
	\centering
	\begin{tabular}{|c||c|c|c|}
		\hline
		  & Iteration 0 & Iteration 1 & Iteration 4\\
		\hline
		\makecell{Ego car \\ $H_{path}$[m]} & 0.589 & 0.206 & 0.146\\\hline
		\makecell{Ego car \\ $H_{velocity}$[m/s]} & 0.799 & 0.373 & 0.309\\ \hline
		\makecell{Ego car \\ $H_{cost}$} & 6.226 & 5.736 & 3.415\\\hline
		\makecell{Ego car \\ Total KPI} & 19.874 & 16.542 & 5.89\\\hline
		\makecell{xDT cars \\ $H_{path}$[m]} & 0.66$\pm$0.296 & 0.209$\pm$0.015 & 0.159$\pm$0.036\\\hline
		\makecell{xDT cars \\ $H_{velocity}$[m/s]} & 0.9432$\pm$0.416 & 0.3699$\pm$0.031 & 0.3178$\pm$0.021\\\hline
	\end{tabular}
\end{table}

\begin{figure}%
	\centering
%
%
\usetikzlibrary{plotmarks}
\definecolor{mycolor1}{rgb}{0.00000,0.44700,0.74100}%
\definecolor{mycolor2}{rgb}{1.00000,0.41000,0.16000}%
\definecolor{mycolor3}{rgb}{0.85000,0.32500,0.09800}%
\begin{tikzpicture}

\begin{axis}[%
width=.7\columnwidth,
height=2.0cm,
at={(0cm,0cm)},
scale only axis,
xmin=-0.5,
xmax=4.5,
xlabel style={font=\color{white!15!black}},
xlabel={Tuning iteration k},
ymin=0,
ymax=6.0,
ylabel={Cost RMS},
axis background/.style={fill=white},
axis x line*=bottom,
axis y line*=left
]
\addplot[ybar, bar width=0.2, fill=black, draw=black, area legend] table[row sep=crcr] {%
-0.15	5.1919975951587\\
};
\addplot[ybar, bar width=0.2, fill=black, draw=black, area legend] table[row sep=crcr] {%
0.85	0.755246167745157\\
};
\addplot[ybar, bar width=0.2, fill=black, draw=black, area legend] table[row sep=crcr] {%
1.85	0.759679880635871\\
};
\addplot[ybar, bar width=0.2, fill=black, draw=black, area legend] table[row sep=crcr] {%
2.85	0.644461383625335\\
};
\addplot[ybar, bar width=0.2, fill=black, draw=black, area legend] table[row sep=crcr] {%
3.85	0.643870899606673\\
};
\end{axis}

\begin{axis}[%
width=.7\columnwidth,
	height=2.0cm,
	at={(0cm,0cm)},
	scale only axis,
	xmin=-0.5,
	xmax=4.5,
	xlabel style={font=\color{white!15!black}},
	xlabel={Tuning iteration k},
	every outer y axis line/.append style={mycolor3},
	every y tick label/.append style={font=\color{mycolor3}},
	every y tick/.append style={mycolor3},
	ymin=0,
	ymax=0.3,
	ylabel style={font=\color{mycolor3}, align=center},
	ylabel={Path tracking RMS},
	axis x line*=bottom,
	axis y line*=right
	]
	\addplot[ybar, bar width=0.2, line width=1.1pt, fill=black, draw=mycolor2, area legend] table[row sep=crcr] {%
		0.15	0.283981356141803\\
	};
	\addplot[ybar, bar width=0.2,line width=1.1pt, fill=black, draw=mycolor2, area legend] table[row sep=crcr] {%
		1.15	0.228887247430915\\
	};
	\addplot[ybar, bar width=0.2,line width=1.1pt, fill=black, draw=mycolor2, area legend] table[row sep=crcr] {%
		2.15	0.180398435874687\\
	};
	\addplot[ybar, bar width=0.2,line width=1.1pt, fill=black, draw=mycolor2, area legend] table[row sep=crcr] {%
		3.15	0.13113868654775\\
	};
	\addplot[ybar, bar width=0.2,line width=1.1pt, fill=black, draw=mycolor2, area legend] table[row sep=crcr] {%
		4.15	0.111068117323187\\
	};
\end{axis}

\begin{axis}[%
width=.7\columnwidth,
height=2cm,
at={(0cm,2.4cm)},
scale only axis,
xmin=-0.5,
xmax=4.5,
ymin=0,
ymax=20,
ylabel style={font=\color{white!15!black}, align=center},
ylabel={Weight on \\ path tracking},
axis background/.style={fill=white},
axis x line*=bottom,
axis y line*=left,
legend style={at={(.01,0.8)},anchor=west},
legend style={draw=none, fill=none}
]
\addplot [color=blue, line width=2.0pt, mark=asterisk, mark options={solid, blue}]
  table[row sep=crcr]{%
0	1\\
1	5.276624679565\\
2	14.077550888062\\
3	18.038549423218\\
4	18.492259979248\\
};
\addlegendentry{Without DR}

\addplot [color=black, line width=2.0pt, mark=asterisk, mark options={solid, black}]
  table[row sep=crcr]{%
0	1\\
1	1.62227165699\\
2	2.140010595322\\
3	3.275551080704\\
4	4.228051662445\\
};
\addlegendentry{With DR}
\end{axis}

\begin{axis}[%
width=.7\columnwidth,
height=4cm,
at={(0in,0in)},
scale only axis,
xmin=0,
xmax=1,
ymin=0,
ymax=1,
axis line style={draw=none},
ticks=none,
axis x line*=bottom,
axis y line*=left
]
\end{axis}
\end{tikzpicture}%
		\caption{Adaptation process: evolution of path tracking and \ac{NMPC} cost RMS on path 1, alongside the path tracking parameter $Q_w$. The learning rate on cost improvement is large, and the performance is enhanced within few iterations} 
		\label{fig:mil_trainingPath1}
\end{figure}
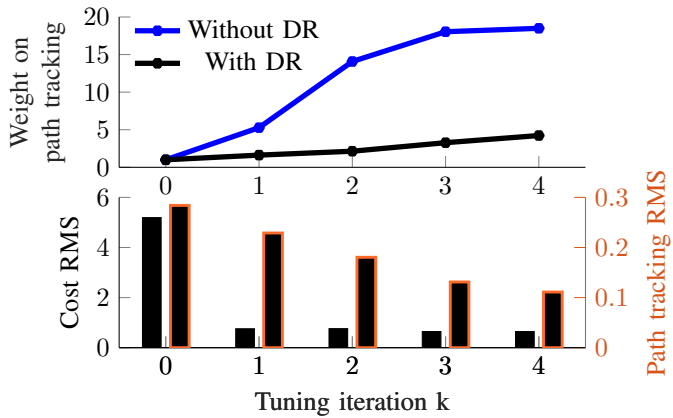

Keeping the training in simulation only, we show the benefit of injecting DR in the form of noise, on the adaptation process as shown in Figure~\ref{fig:mil_trainingPath1}. Due to the awareness that the target domain contains input and output noise, we inject the sample \ac{xDT}s with noise. This helps to avoid overfitting the trained controller to a noiseless environment. From Figure~\ref{fig:mil_trainingPath1}, one can see that in the presence of noise, the parameter update on path tracking is more conservative than in the noiseless case. This is due to the higher weight on the path deviation that causes the controller to react rapidly to path error deviation which are corrupted with noise. Therefore, it is clear that the controller adapts to the present uncertainties directly without estimating them by sampling the 19 parallel \ac{xDT}s.
The training process is shown in the lower plot of Figure~\ref{fig:mil_trainingPath1}. The importance of the method resides in its sampling and data efficiency, as it takes few iterations to reach an acceptable performance level. The \ac{NMPC} cost immediately drops to a plateau from the first iteration with an improvement of 87\%. The tracking performance improves by 60\%  in just 4 iterations. %

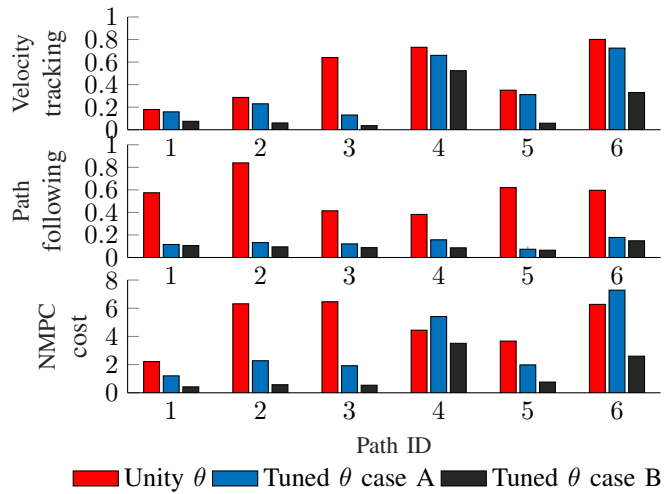
\begin{figure}%
	\centering
%
%
\definecolor{mycolor1}{rgb}{0.00000,0.44706,0.74118}%
\definecolor{mycolor2}{rgb}{0.14902,0.14902,0.14902}%
\begin{tikzpicture}

\begin{axis}[%
width=.8\columnwidth,
height=1.5cm,
at={(0cm, 0cm)},
scale only axis,
legend style={at={(axis cs:-0.2,-4.5)},anchor=north west},
legend style={draw=none, legend columns=-1},
xmin=0.511111111111111,
xmax=6.48888888888889,
xlabel style={font=\color{white!15!black}},
xlabel={\small Path ID},
ymin=0,
ymax=8,
ylabel style={font=\color{white!15!black}, align=center},
ylabel={\small NMPC\\cost},
axis background/.style={fill=white},
axis x line*=bottom,
axis y line*=left
]
\addplot[ybar, bar width=0.178, fill=red, draw=black, area legend] table[row sep=crcr] {%
0.78	2.21298877779509\\
};
\addplot[ybar, bar width=0.178, fill=mycolor1, draw=black, area legend] table[row sep=crcr] {%
1	1.20167409172908\\
};
\addplot[ybar, bar width=0.178, fill=mycolor2, draw=black, area legend] table[row sep=crcr] {%
1.22	0.421177344927018\\
};
\addplot[ybar, bar width=0.178, fill=red, draw=black, area legend] table[row sep=crcr] {%
1.78	6.31311185131318\\
};
\addplot[ybar, bar width=0.178, fill=mycolor1, draw=black, area legend] table[row sep=crcr] {%
2	2.27510867741299\\
};
\addplot[ybar, bar width=0.178, fill=mycolor2, draw=black, area legend] table[row sep=crcr] {%
2.22	0.572534084377998\\
};
\addplot[ybar, bar width=0.178, fill=red, draw=black, area legend] table[row sep=crcr] {%
2.78	6.460588889239\\
};
\addplot[ybar, bar width=0.178, fill=mycolor1, draw=black, area legend] table[row sep=crcr] {%
3	1.91929224780534\\
};
\addplot[ybar, bar width=0.178, fill=mycolor2, draw=black, area legend] table[row sep=crcr] {%
3.22	0.536368920548486\\
};
\addplot[ybar, bar width=0.178, fill=red, draw=black, area legend] table[row sep=crcr] {%
3.78	4.44137662113725\\
};
\addplot[ybar, bar width=0.178, fill=mycolor1, draw=black, area legend] table[row sep=crcr] {%
4	5.40931642347297\\
};
\addplot[ybar, bar width=0.178, fill=mycolor2, draw=black, area legend] table[row sep=crcr] {%
4.22	3.50964768808024\\
};
\addplot[ybar, bar width=0.178, fill=red, draw=black, area legend] table[row sep=crcr] {%
4.78	3.6673410905317\\
};
\addplot[ybar, bar width=0.178, fill=mycolor1, draw=black, area legend] table[row sep=crcr] {%
5	1.9795076364982\\
};
\addplot[ybar, bar width=0.178, fill=mycolor2, draw=black, area legend] table[row sep=crcr] {%
5.22	0.761187544436039\\
};
\addplot[ybar, bar width=0.178, fill=red, draw=black, area legend] table[row sep=crcr] {%
5.78	6.27292305599747\\
};
\addplot[ybar, bar width=0.178, fill=mycolor1, draw=black, area legend] table[row sep=crcr] {%
6	7.28221210479739\\
};
\addplot[ybar, bar width=0.178, fill=mycolor2, draw=black, area legend] table[row sep=crcr] {%
6.22	2.59923927953253\\
};
\addlegendentry{Unity $\theta$}
\addlegendentry{Tuned $\theta$ case A}
\addlegendentry{Tuned $\theta$ case B}
\end{axis}

\begin{axis}[%
width=.8\columnwidth,
height=1.5cm,
at={(0cm,1.8cm)},
scale only axis,
xmin=0.511111111111111,
xmax=6.48888888888889,
ymin=0,
ymax=1,
ylabel style={font=\color{white!15!black}, align=center},
ylabel={\small Path\\following},
axis background/.style={fill=white},
axis x line*=bottom,
axis y line*=left
]
\addplot[ybar, bar width=0.178, fill=red, draw=black, area legend] table[row sep=crcr] {%
0.78	0.574117286886797\\
};
\addplot[ybar, bar width=0.178, fill=mycolor1, draw=black, area legend] table[row sep=crcr] {%
1	0.115250872147308\\
};
\addplot[ybar, bar width=0.178, fill=mycolor2, draw=black, area legend] table[row sep=crcr] {%
1.22	0.106154880579449\\
};
\addplot[ybar, bar width=0.178, fill=red, draw=black, area legend] table[row sep=crcr] {%
1.78	0.839162054703727\\
};
\addplot[ybar, bar width=0.178, fill=mycolor1, draw=black, area legend] table[row sep=crcr] {%
2	0.13200816289956\\
};
\addplot[ybar, bar width=0.178, fill=mycolor2, draw=black, area legend] table[row sep=crcr] {%
2.22	0.0944401141459108\\
};
\addplot[ybar, bar width=0.178, fill=red, draw=black, area legend] table[row sep=crcr] {%
2.78	0.414144168756051\\
};
\addplot[ybar, bar width=0.178, fill=mycolor1, draw=black, area legend] table[row sep=crcr] {%
3	0.120658482068265\\
};
\addplot[ybar, bar width=0.178, fill=mycolor2, draw=black, area legend] table[row sep=crcr] {%
3.22	0.087570882585009\\
};
\addplot[ybar, bar width=0.178, fill=red, draw=black, area legend] table[row sep=crcr] {%
3.78	0.381984666237667\\
};
\addplot[ybar, bar width=0.178, fill=mycolor1, draw=black, area legend] table[row sep=crcr] {%
4	0.15688508108353\\
};
\addplot[ybar, bar width=0.178, fill=mycolor2, draw=black, area legend] table[row sep=crcr] {%
4.22	0.0858351206564038\\
};
\addplot[ybar, bar width=0.178, fill=red, draw=black, area legend] table[row sep=crcr] {%
4.78	0.61997216494578\\
};
\addplot[ybar, bar width=0.178, fill=mycolor1, draw=black, area legend] table[row sep=crcr] {%
5	0.0730576378623635\\
};
\addplot[ybar, bar width=0.178, fill=mycolor2, draw=black, area legend] table[row sep=crcr] {%
5.22	0.0646442771929131\\
};
\addplot[ybar, bar width=0.178, fill=red, draw=black, area legend] table[row sep=crcr] {%
5.78	0.595968910131049\\
};
\addplot[ybar, bar width=0.178, fill=mycolor1, draw=black, area legend] table[row sep=crcr] {%
6	0.17780274290233\\
};
\addplot[ybar, bar width=0.178, fill=mycolor2, draw=black, area legend] table[row sep=crcr] {%
6.22	0.148095214956322\\
};
\end{axis}

\begin{axis}[%
width=.8\columnwidth,
height=1.5cm,
at={(0cm,3.5cm)},
scale only axis,
xmin=0.511111111111111,
xmax=6.48888888888889,
ymin=0,
ymax=1,
ylabel style={font=\color{white!15!black}, align=center},
ylabel={\small Velocity\\tracking},
axis background/.style={fill=white},
axis x line*=bottom,
axis y line*=left
]
\addplot[ybar, bar width=0.178, fill=red, draw=black, area legend] table[row sep=crcr] {%
0.78	0.179350231109659\\
};
\addplot[ybar, bar width=0.178, fill=mycolor1, draw=black, area legend] table[row sep=crcr] {%
1	0.158413142863973\\
};
\addplot[ybar, bar width=0.178, fill=mycolor2, draw=black, area legend] table[row sep=crcr] {%
1.22	0.0746078223589789\\
};
\addplot[ybar, bar width=0.178, fill=red, draw=black, area legend] table[row sep=crcr] {%
1.78	0.286386002333794\\
};
\addplot[ybar, bar width=0.178, fill=mycolor1, draw=black, area legend] table[row sep=crcr] {%
2	0.229834477333556\\
};
\addplot[ybar, bar width=0.178, fill=mycolor2, draw=black, area legend] table[row sep=crcr] {%
2.22	0.060109728522296\\
};
\addplot[ybar, bar width=0.178, fill=red, draw=black, area legend] table[row sep=crcr] {%
2.78	0.640391506038165\\
};
\addplot[ybar, bar width=0.178, fill=mycolor1, draw=black, area legend] table[row sep=crcr] {%
3	0.130480887289139\\
};
\addplot[ybar, bar width=0.178, fill=mycolor2, draw=black, area legend] table[row sep=crcr] {%
3.22	0.0360164978543556\\
};
\addplot[ybar, bar width=0.178, fill=red, draw=black, area legend] table[row sep=crcr] {%
3.78	0.731158147305015\\
};
\addplot[ybar, bar width=0.178, fill=mycolor1, draw=black, area legend] table[row sep=crcr] {%
4	0.659680438504496\\
};
\addplot[ybar, bar width=0.178, fill=mycolor2, draw=black, area legend] table[row sep=crcr] {%
4.22	0.52353689011641\\
};
\addplot[ybar, bar width=0.178, fill=red, draw=black, area legend] table[row sep=crcr] {%
4.78	0.350514192312324\\
};
\addplot[ybar, bar width=0.178, fill=mycolor1, draw=black, area legend] table[row sep=crcr] {%
5	0.311503645936705\\
};
\addplot[ybar, bar width=0.178, fill=mycolor2, draw=black, area legend] table[row sep=crcr] {%
5.22	0.0577462728970389\\
};
\addplot[ybar, bar width=0.178, fill=red, draw=black, area legend] table[row sep=crcr] {%
5.78	0.800976251311321\\
};
\addplot[ybar, bar width=0.178, fill=mycolor1, draw=black, area legend] table[row sep=crcr] {%
6	0.72380386135624\\
};
\addplot[ybar, bar width=0.178, fill=mycolor2, draw=black, area legend] table[row sep=crcr] {%
6.22	0.330136043346561\\
};
\end{axis}

\begin{axis}[%
width=.8\columnwidth,
height=5cm,
at={(0in,0in)},
scale only axis,
xmin=0,
xmax=1,
ymin=0,
ymax=1,
axis line style={draw=none},
ticks=none,
axis x line*=bottom,
axis y line*=left
]
\end{axis}
\end{tikzpicture}%
	\caption{Validation process: RMS for the closed-loop \ac{NMPC} cost, velocity tracking and path following errors over the different paths in the library. Red: unit $\theta$. Blue: tuned $\theta$ only on path \#1. Black: tuned $\theta$ with path domain randomization} 
	\label{fig:mil_validation_allpathsCL}
\end{figure}

To determine the scalability of the method, we consider several paths with different driving styles (aggressive human driving, smooth optimal path) and different curvatures. \textcolor{responsecolor}{From a random set of initial position and target parking spots in the parking lot, an A* hybrid planner generates a library of reference paths, with different velocity profiles, different curvatures (sharp turns, straight lines, sinusoidal driving). Some of those paths will lead the car to drive on the uphill part of the parking lot, and others will trigger an aggressive driving style. For the sake of clarity, we focus on 6 of those chosen paths to study the performance transferability between varying paths}. \textcolor{responsecolor}{In comparison with the TiL work in~\cite{twin_in_the_loop_23}, the objective is not to tune the DT to capture the real world dynamics for every path on its own, but rather generate a controller that is robust enough to a variation of scenarios}. Let \textit{case A} be the result of tuning using only the first path. When applied to the other unseen trajectories, the trained controller is generally better in velocity and path tracking than the unit $\theta$ initial weight matrix, as in Figure~\ref{fig:mil_validation_allpathsCL}. However, on dynamic trajectories (path 4 and 6), the controller results in a higher cost and computational burden. This indicates that the controller failed to generalize to other trajectories by over-tuning for path 1 only. Therefore, we include DR on the path by using 80\% of the trajectories' dataset for training and generate the tuned matrix of \textit{case B} presented in Figure~\ref{fig:mil_validation_allpathsCL}. This second tuned matrix generalizes better to other tasks by finding a local optima for the \ac{NMPC} parameters improving all the performance metrics. The overall RMS of the \ac{NMPC} cost time-series drops from 5.1481 (unity) to 4.0107 (case A) to 1.846 (case B). For path following, the RMS on the tracking error drops from 0.59031 (unity) to 0.13344 (case A) to 0.10111 (case B). Finally, the RMS on velocity tracking drops from 0.55126 (unity) to 0.438 (case A) to 0.25719 (case B).%

\begin{figure}%
	\centering
	\input{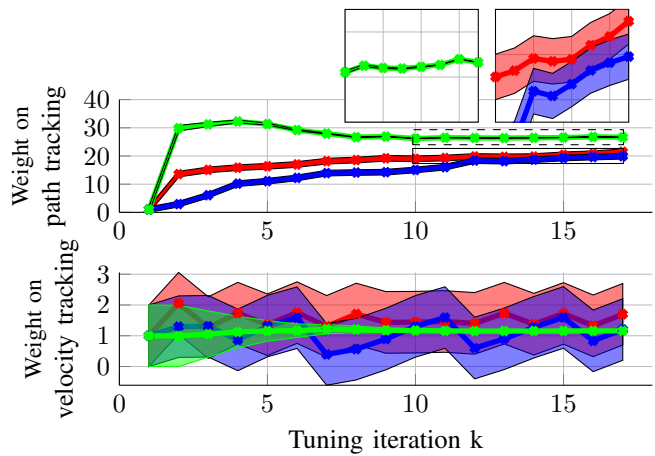}
	\caption{Impact of adaptive covariance matrices proposed in this work (green), in comparison with~\cite{ALLAMAA2022385} (red) and~\cite{menner2023automated} (blue). The shaded areas represent the uncertainties on the parameter} 
	\label{fig:adaptive_UKFwSPSA_performance_Q_evo}
\end{figure}

Finally, we benchmark our proposed approach against~\cite{menner2023automated} and our previous work~\cite{ALLAMAA2022385} where the covariance matrices were kept constant. It is worth noting that tuning the covariance matrices is by itself a bottleneck of this approach and affects the results directly. In Figure~\ref{fig:adaptive_UKFwSPSA_performance_Q_evo}, we perform a tuning campaign where only two parameters are tuned: the weights on path deviation and velocity tracking. The work of~\cite{menner2023automated} is outperformed by~\cite{ALLAMAA2022385} in terms of adaptation speed given the inclusion of the SPSA step. However, a closer look at the parameter posterior covariance indicates that both approaches lead to a relatively large uncertainty in the parameter set. This causes unnecessary sampling of the \ac{xDT}s even after the performance index has converged. In comparison, by including the adaptive covariance as in $\DCAUKS$, the guided search directions are more reflective of the true sensitivity of the parameters with respect to the real-world performance. This leads to a faster convergence of the method, in terms of both step size and sampling direction. 

\subsection{Stochastic black-box optimization}
\begin{figure}%
	\centering
	\input{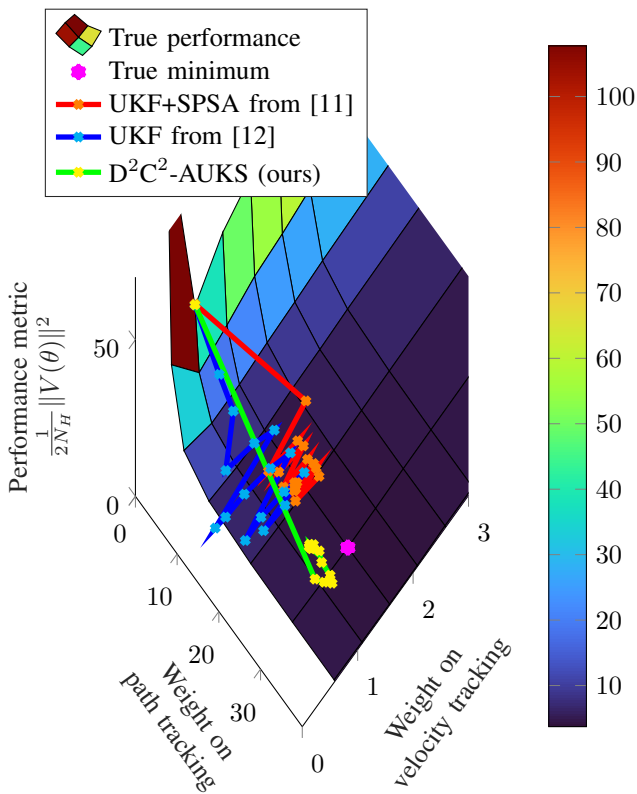}
	\caption{Iterative adaptation framework and comparison with the true performance metric \ac{KPI} on a fine grid} 
	\label{fig:3dplot_performance_Q_evo_zoomed_in}
\end{figure}

The proposed framework helps to simultaneously optimize the controller parameters, for a given user defined performance. As the adaptation takes place in the target domain in a closed-loop fashion, under several sources of noise, delays, and uncertainties, we seek to validate that the algorithm does indeed solve iteratively an optimization problem. Therefore, we run oracles with randomization in the high-fidelity simulator, each with a varying combination of the parameters to tune $\theta = [\text{diag}(Q_\theta), \text{diag}(R_\theta)]$. We collect for each scenario the total performance metric $\frac{1}{2N_T}\lVert V(\theta)\rVert ^2$ which is proportional to from~\eqref{eq:highlevel_bilevelOpt}. For the sake of visualization, we consider the two most dominant parameters: the weight on path tracking error $Q_w$ and the velocity tracking error $Q_{vx}$. The results of a parameter adaptation campaign are shown in Figure~\ref{fig:3dplot_performance_Q_evo_zoomed_in}. Starting from a unit set of parameters, our $\DCAUKS$ algorithm reaches a minimal cost within few iterations, as the parameters converge and the uncertainties shrink. In comparison with~\cite{menner2023automated} and~\cite{ALLAMAA2022385} which employ constant noise covariance matrices, the proposed approach exhibits no oscillation around the optimal set of parameters. $\DCAUKS$ benefits from the data driven adaptation of the parameter perturbation and noise covariance matrices $C_{\Delta\theta}, C_v$. In fact, the two covariance matrices adapt to the actual noise level in the target domain through uncertainties on the parameters and on the Sim2Real gap respectively. Thus, the algorithm compensates for those uncertainties.

\begin{figure}%
	\centering
	\input{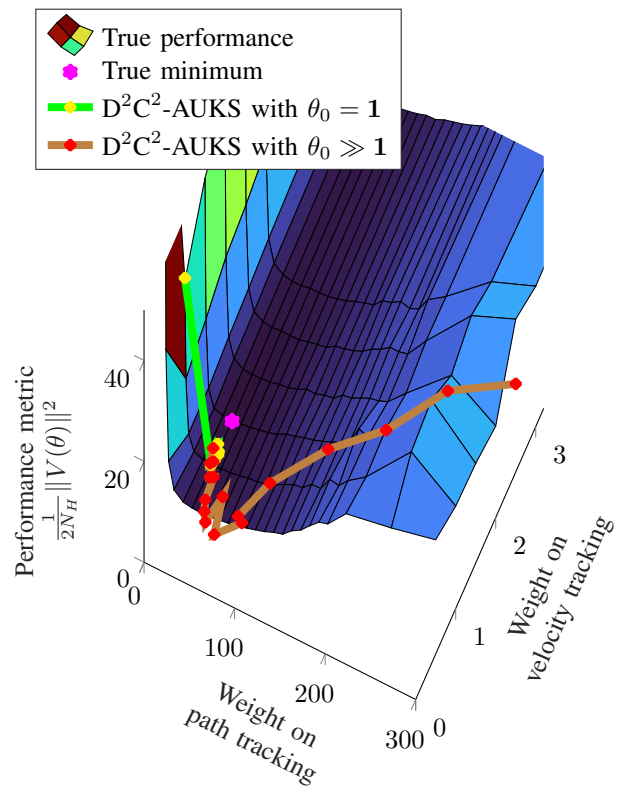}
	\caption{Iterative adaptation framework solving for the stochastic black-box optimization problem in~\eqref{eq:highlevel_bilevelOpt}. For an initial $\theta_0 \gg \mathbf{1}$, the iterative framework recovers from the local minima and reach the optimal \ac{KPI} value} 
	\label{fig:3dplot_performance_Q_evo}
\end{figure}

Moreover, we initiate the parameter set in the neighborhood of a local minimum as seen in Figure~\ref{fig:3dplot_performance_Q_evo}, with $ \theta_0 \gg \mathbf{1}$, indicating that every element of $\theta$ is larger than 1. This is in comparison to the case with $\theta=\mathbf{1}$ where the individual elements of $\theta$ are equal to 1. Starting from $Q_w = 300, Q_{vx} = 3$, the algorithm escapes the local minimum by exploring the local behavior of the \ac{KPI}, and converges to the minimum value of the total performance metric, or equivalently reaches the maximum performance. This is due to the explorative nature of the framework in sampling several parallel \ac{xDT}s in order to form the derivative-free update and moving along the average performance improving direction.
%

\section{Sim2Real experimental validation with real-time automatic adaptation}
\label{section:adaptation_experimental}
One of the novelties in this work resides in the experimental validation of the adaptation and parameter tuning on road vehicles. Our platform is a SimRod drive-by-wire vehicle as seen in Figure~\ref{fig:sim2real_vil_prob}. The \ac{NMPC} runs on a dSPACE MicroAutobox III with a real-time operating system for embedded applications. The \ac{NMPC} is sampled at $20\,\textrm{Hz}$ as a low-level controller, commanding the steering and acceleration/deceleration rates. This section presents the results of the Sim2Real automatic \ac{NMPC} adaptation using $\DCAUKS$. 

\subsection{Importance of Sim2Real transfer}
First, the \ac{NMPC} calibrated in simulation is transferred to the real-world, and the closed-loop performance is shown in Figure~\ref{fig:sim2real_vil_prob}. It is a direct observation, that the \ac{NMPC} parameters were over fitted in simulation to generate an optimal performance level in \ac{NMPC} cost and tracking errors, which motivates our work. Similar to most applications, including learning and classical control methods, engineers employ this Sim2Real methodology that consists of parameter tuning in simulation until a satisfying performance is reached. However, this approach often fails to transfer to the real world. This raises the need for 1) an adaptive scheme that can learn from the data collected in the real world 2) closing the loop between the prior expectation (simulation) and the actual (real-world) closed-loop performance 3) a method to rapidly retune the parameters and avoid manual tuning.

On the right-hand side of Figure~\ref{fig:sim2real_vil_prob}, we present an alternative methodology for the Sim2Real transfer. The \ac{xDT}s are spawned in parallel with the real vehicle, to smartly explore the parameters to tune and randomize the disturbances $\xi$. Exploration takes place rapidly and safely so that the parameters update on the target vehicle is facilitated. The performance of our approach is validated, and the results are shown in the Figure~\ref{fig:sim2real_vil_prob} and Table~\ref{tab:vil_performance_autotuned}. The path and parking tracking errors dropped to below 30 cm and 15 cm respectively, and the \ac{NMPC} cost is minimal. We start the training with $\theta=\mathbf{1}$, which could represent a case of complete absence of engineering expertise. The closed-loop performance is shown in orange. After 4 iterations (purple curve), lasting a total of approximately 10 minutes including the repositioning time, the \ac{NMPC} is tuned to optimize for this target configuration that is corrupted with steering actuator delays. The performance improvement is significant, and the end-of-line tuning time is cut down from hours to just few minutes.



\subsection{Generalization to different paths}
\begin{figure}
	\centering
	\includegraphics[width=0.45\textwidth, height=6.5cm]{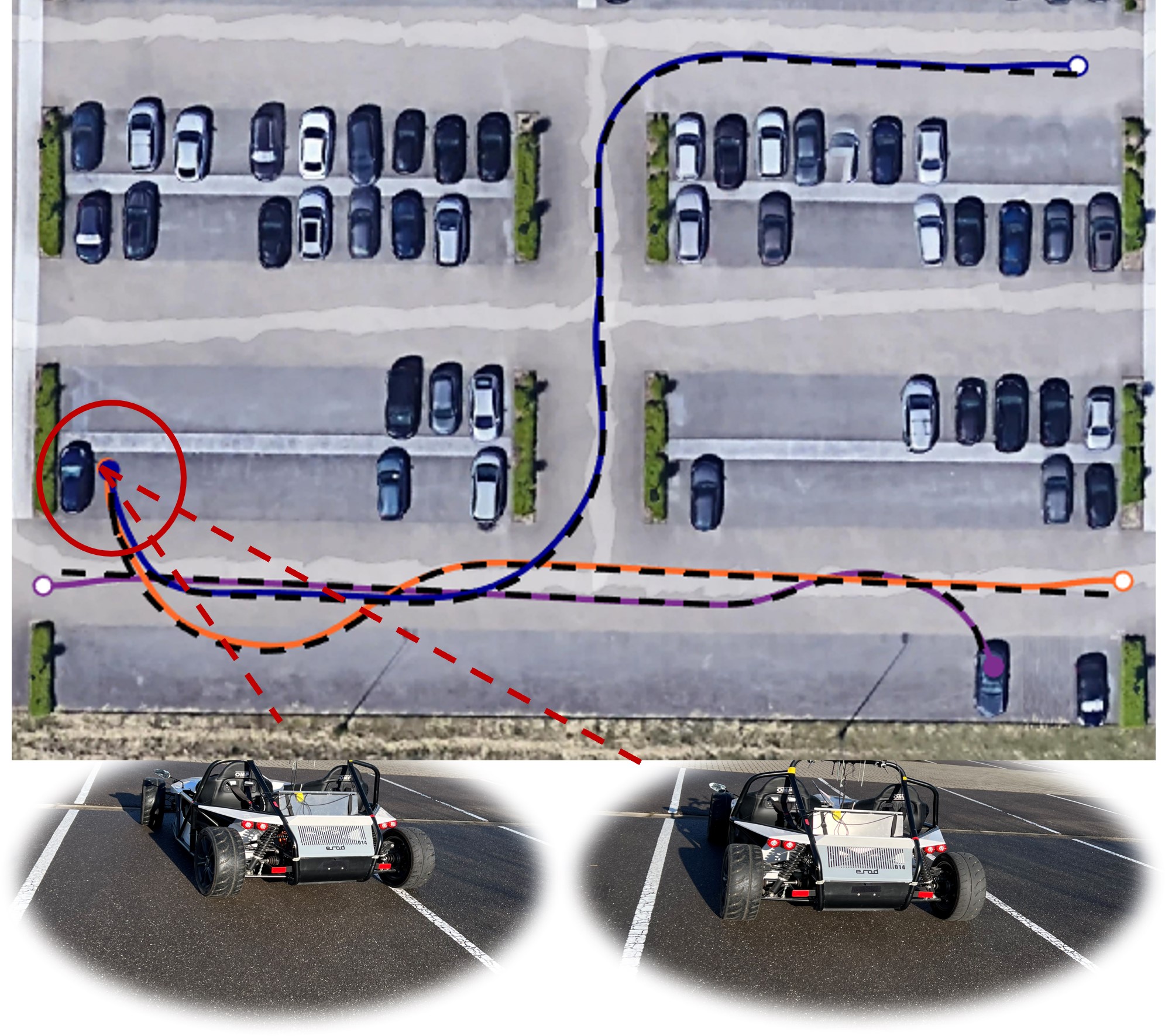}
	\caption{ViL: policy trained by in closed-loop by randomizing over the paths. Reference path (dashed line), closed-loop ego position (colored). The controller is tuned to park properly (right) in few iterations ($\sim$10 minutes) after starting from a suboptimal performance (left)}
	\label{fig:sim2real_vil_severalpaths}
\end{figure}

Depending on the intended application, the presented algorithm could be used to either reach an optimal parameter tuning for one specific task and environment condition, or to generalize for a set of tasks and conditions. In other words, if the vehicle is to repeatedly follow the same path, then a task specific parameter optimization could be reached. However, if the vehicle is set to drive on different curvatures, with varying velocity profiles and road conditions, then $\DCAUKS$ is employed to adapt the parameters such that the performance is conserved across the tasks. To tackle the latter case, we train our algorithm starting from a unit $\theta=\mathbf{1}$ with DR on the chosen path and the vehicle's initial conditions. This helps avoiding to overfit for one specific task. The \ac{ViL}, real-world adaptation of a \ac{RTNMPC} is shown in Figure~\ref{fig:sim2real_vil_severalpaths} and Table~\ref{tab:vil_performance_autotuned}. In particular, the blue path, encounters a road grade of approximately $4\%$. By tuning on the orange path solely, then validating on the blue path, the vehicle fails to climb the slope. This is due a to a calibrated controller with low weight on the velocity tracking component in the \ac{NMPC} cost $Q_{vx}$. However, we combine trajectories of different driving styles (human driven and smooth A* planner generated), with a multitude of target parking spots and perturbed initial conditions. Thus, the learning-based optimal controller adaptation generalizes for a variety of scenarios with improved performance overall. Three examples of tracked paths are shown in Figure~\ref{fig:sim2real_vil_severalpaths}, where the real vehicle is controlled to park within $15\,\textrm{cm}$ of accuracy using a RTNMPC as in Figure~\ref{fig:sim2real_vil_severalpaths}. The total RMS of the path tracking error, velocity tracking error and closed \ac{NMPC} cost over a multitude of trajectories are $0.295\,\textrm{m}$, $0.642\,\textrm{m/s}$, and $2.417$ respectively as represented in Table~\ref{tab:vil_performance_autotuned}.

\begin{table}[hb!] 
	\begin{center}
		\caption{\textcolor{responsecolor}{RMS of the individual outputs,Sim2Real performance degradation and recovery with automatic controller calibration over different paths from Figures~\ref{fig:sim2real_vil_prob} and~\ref{fig:sim2real_vil_severalpaths} \label{tab:vil_performance_autotuned}}}
		\begin{tabular}{|c||c|c|c|c|c|c}
			\hline
			Tuning & \makecell{Tracking \\ error [m]} & \makecell{Velocity \\error [m/s]} & \makecell{NMPC \\cost} & \makecell{Total \\ KPI} & \makecell
			{Sim2Real \\ Deterioration} \\\hline
			SimOptimal in Sim & 0.1327 & 0.2148 & 2.4645 & 3.0678 & 0\% \\\hline
			SimOptimal in Real & 1.2015 & 0.8587 & 73.4097 & 2.69e3  & 8.77e4\%\\ \hline
			Unit $\theta$ in Real & 1.365 & 0.39133 & 8.933 & 40.906 & 1.21e3\%\\\hline
			\thead{Autotuned in Real} & \thead{0.295} & \thead{0.642} & \thead{2.417} & \thead{3.17} & \thead{3.34\%}\\\hline
		\end{tabular}
	\end{center}
\end{table}

\textcolor{responsecolor}{The necessity of the automatic calibration is to recover the expected performance from simulation, when transferring to the real-world. That is, the parameter search and optimization seek to directly compensate for model mismatches and noises with the single objective of optimizing the performance in~\eqref{eq:highlevel_bilevelOpt}. For this reason, we summarize the results of the method in~\ref{tab:vil_performance_autotuned}: the total performance multi-objective $V$ is composed of the path tracking error $Y_{path}$, the velocity tracking error $Y_{velocity}$ and the closed-loop \ac{NMPC} cost $Y_{cost}$. They are measured online on the real system, and we report their RMS over the time window $T$. We compare three tuning combinations to a controller tuned perfectly in simulation (\textit{SimOptimal in Sim}). First, \textit{SimOptimal in Real} is the evaluation of the optimal tuning obtained in simulation that is transferred to the real world. This is the typical case of a one-directional XiL testing with a controller that is over optimized in simulation, as reported in the left part of Figure~\ref{fig:sim2real_vil_prob}. Second, controller tuning \textit{Unit in Real} is an initial uneducated guess on the parameters with $\theta = \textbf{1}$. It serves as a starting point for our automatic calibration, visualized in orange in Figure~\ref{fig:sim2real_vil_prob}. Third, the last controller tuning \textit{Autotune in Real} is automatically tuned in the real-world using $\DCAUKS$, as reported in the purple plot in Figure~\ref{fig:sim2real_vil_prob} and the closed-loop trajectories of Figure~\ref{fig:sim2real_vil_severalpaths}. \textit{SimOptimal in Real} suffers from a catastrophic performance degradation that is 8.77e4\% higher than the one expected in simulation. All components of the \ac{KPI} are worsened, specifically the tracking error and \ac{NMPC} cost. This is caused by the delays in the low-level actuation and a high weight on the tracking error that causes the controller to fail in tracking the trajectories. Next, for \textit{Unit in Real}, a similar bad performance in terms of path tracking is noticed, with an RMS of $1.37\,\textrm{m}$ because of the low weight on the tracking error. Most importantly with \textit{Autotuned in Real}, in under 10 minutes we are able to automatically calibrate the controller on-the-go. By feeding with the real-world data, the expected performance from simulation is recovered with little testing effort on the target system. That is, we recover the total \ac{KPI} value of $3.17$, which is 3.34\% higher than the simulated performance, showcasing the ability to close the Sim2Real gap. However, as the individual metrics contributing to $V$ are optimized simultaneously, and given the complex nature of the optimization problem, the tracking errors and the \ac{NMPC} cost are not identically matched to their simulated performances. Nevertheless, the path tracking error is enhanced by dropping from $1.365\,\textrm{m}$ down to $0.295\,\textrm{m}$ over the multitude of trajectories. This accounts for the noise in estimation, as well as the initial condition of the vehicle that was off the path. In terms of energy, the \ac{NMPC} cost is dropped by a factor of $3.7$ from $8.933$ down to $2.417$, in 4 iterations.}

\section{Limitations}
\label{section:limitations}
The presented framework is to be used as a closed-loop black-box optimizer for the performance of a parametrizable controller. The following limitations are to be noted:
\textcolor{responsecolor}{
	\begin{itemize}
		\item Further extensions include tuning the constraints of the \ac{NMPC} such that the real system meets those constraints, or tuning the prediction model in the \ac{NMPC} to close the Sim2Real gap on the output responses. However, only the work concerning the automatic and on-the-fly calibration of a parametrized cost function in an \ac{NMPC} for real-world \ac{AD} has been validated.
		\item We assume that the employed \ac{NMPC} under the hood maintains the safety requirements and acts as a safety filter for the data-driven automatic calibrator. The usage of \ac{xDT}s to explore the control parameter space adds layers of safety and speed-up, as we not all the variation of parameters are sampled on the real system to form the Kalman gain and stochastic gradient steps.
		\item We do not provide any guarantees on the effectiveness of combining SPSA with \ac{UKF} for a derivative-free optimization. However, the necessary convergence guarantees with diminishing step size and exploration step size, support the claim that the overall adaptation framework is at least converging to a local minimum.
		\item Given the highly non-convex nature of the optimization problem, global optimum might not be reached, and it is possible to settle on a local minimum as we optimize by sampling several \ac{xDT}s simultaneously and step in the direction of average expected improvement. This work is characterized by an automatic calibrator that improves the performance, with the least amount of real-world sampling and maximal digital world exploration, in a limited amount of time.
	\end{itemize}
	}

\section{Conclusion}
\label{section:conclusion}
In this work, we have presented a learning-based adaptation scheme for parametric controllers that allows a faster, safer, and cheaper transfer from one domain to another, through exploration in the executable digital twin domain and exploitation in the target domain. In particular, we validate the methodology on a parameter tuning and adaptation for a real-time \ac{NMPC} controller in an autonomous valet parking framework. The proposed method is data and sampling efficient, and directly optimizes the closed-loop performance in the target domain with few iterations. We combine iterative Kalman filtering techniques with SPSA to solve a derivative-free optimization problem with possibly non-differentiable objectives. Experimental validation shows that the \ac{NMPC} can be tuned in the matter of few minutes to significantly improve performance and reduce the incurred \ac{NMPC} cost, while generalizing for different tasks. Moreover, we recover a Sim2Real gap of almost a factor of 1 through careful combination of simulated exploration and real data exploitation.
The method is sample efficient as it gathers useful information with fewer interactions with the target system, thus reducing the spent time and resources spent on end-of-line tuning. 

\section*{Acknowledgments}
This project has received funding from the European Union’s Horizon 2020 research and innovation programme under the Marie Skłodowska-Curie grant agreement ELO-X No 953348.

\bibliographystyle{IEEEtran}
\bibliography{biblio23lbaNMPC} 

\begin{IEEEbiography}[{\includegraphics[width=1in,height=1.25in,clip]{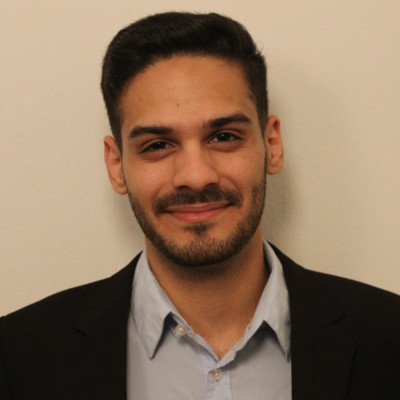}}]{Jean Pierre Allamaa}
	obtained the B.Eng. degree in Mechanical Engineering from the American University of Beirut, Beirut, Lebanon and the M.Sc. degree in Mechanical Engineering with a specialization in Automatic and Control from École Polytechnique Fédérale de Lausanne (EPFL), Lausanne, Switzerland in 2018 and 2020 respectively.  He is currently a Marie-Curie fellow, working towards his industrial Ph.D. in Electrical Engineering at Siemens and in collaboration with the Department of Electrical Engineering (ESAT) at KU Leuven, Leuven, Belgium. 
	
	His research focuses on embedded model predictive control and its intersection with learning, in particular with applications in motion planning and control for autonomous driving and advanced driver assistance systems, in virtual and real environments.
\end{IEEEbiography}

\begin{IEEEbiography}[{\includegraphics[width=1in,height=1.25in,clip,keepaspectratio]{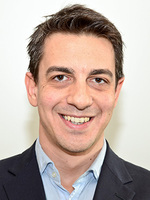}}]{Panagiotis (Panos) Patrinos}
received the
M.Eng. degree in chemical engineering, M.S.
degree in applied mathematics, and the Ph.D.
degree in control and optimization from the National Technical University of Athens, Athens,
Greece, in 2003, 2005, and 2010, respectively.
He is currently an Associate Professor
with the Department of Electrical Engineering
(ESAT), KU Leuven, Leuven, Belgium. In 2014,
he was a Visiting Professor with Stanford University, Stanford, CA, USA. After his Ph.D., he
was a Postdoc with the University of Trento, Trento, Italy, and IMT
Lucca, Lucca, Italy, where he became an Assistant Professor in 2012.
His current research interests include the intersection of optimization,
control and learning, theory and algorithms for structured nonconvex
optimization, and learning-based, model predictive control with a wide
range of applications including autonomous vehicles, machine learning,
and signal processing.

Dr. Patrinos was the corecipient of the 2020 Best Paper Award in the
International Journal of Circuit Theory and Applications.
\end{IEEEbiography}

\begin{IEEEbiography}[{\includegraphics[width=1in,height=1.25in,clip]{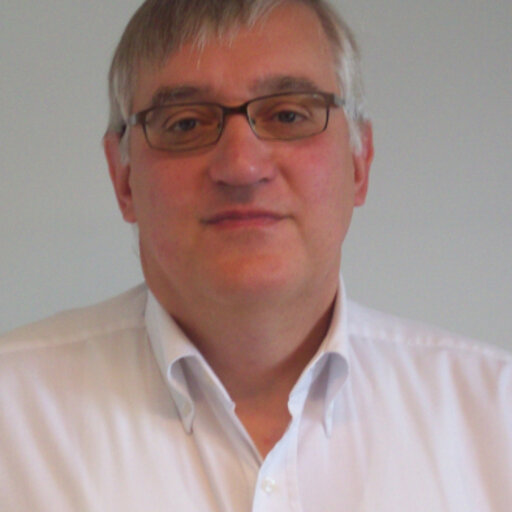}}]{Herman Van der Auweraer}
received the M.Sc. degree in electronic engineering (1980) and the Ph.D. degree in engineering science (1987) from the KU Leuven, Belgium. In 1986, he joined LMS International, Leuven, one of the earliest KU Leuven spin-offs, developing advanced testing and simulation tools for mechatronic product design engineering. LMS became part of Siemens in 2013. His research focus is acoustics, sound quality, and system identification. He was Director Research and Technology Innovation until his retirement in 2023. He continues to support the company’s innovation strategy as Senior Advisor. Furthermore, he is affiliated to KU Leuven as guest professor.
\end{IEEEbiography}

\begin{IEEEbiography}[{\includegraphics[width=1in,height=1.25in,clip]{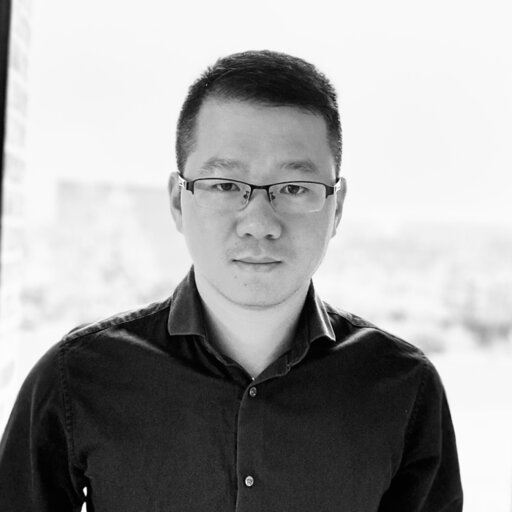}}]{Tong Duy Son}
	received the PhD as a Marie-Curie fellow from Department of Mechanical Engineering, KU Leuven, Belgium in 2016. Since then he is a senior researcher and an R\&D ADAS Manager at Siemens Digital Industries Software, active in European Union and Belgian research programs and supervision of industrial PhDs. His research focuses on autonomous driving control, AI algorithms development, testing and validation methodologies in both virtual and physical environment. 
	
	Dr. Tong was the recipient of the Siemens DF PL Invention of the Year Award.
\end{IEEEbiography}

\end{document}